\documentclass{article}

\usepackage[main, final]{neurips_2026}

\usepackage[utf8]{inputenc} % allow utf-8 input
\usepackage[T1]{fontenc}    % use 8-bit T1 fonts
\usepackage{hyperref}       % hyperlinks
\hypersetup{colorlinks=true,linkcolor=blue,citecolor=blue,urlcolor=blue}
\usepackage{url}            % simple URL typesetting
\usepackage{booktabs}       % professional-quality tables
\usepackage{amsfonts}       % blackboard math symbols
\usepackage{amsmath}       % 
\usepackage{nicefrac}       % compact symbols for 1/2, etc.
\usepackage{microtype}      % microtypography
\usepackage{xcolor}         % colors
\usepackage{xspace}
\usepackage{graphicx}
\usepackage{multirow}
\usepackage{makecell}
\usepackage{mathtools}
\usepackage{fontawesome5}
\usepackage{subcaption} % for subfigs
\makeatletter
\renewcommand{\p@subfigure}{}
\makeatother

\usepackage{caption}
\usepackage{wrapfig} % for intro plot

\newcommand{\methodname}{\textsc{ShadowCLIP}\xspace}

\title{Displacement Geometry Captures Platonic Shared Reality Across Models and Modalities}

\author{
\parbox[t]{0.5\textwidth}{
  \centering
  Chenming Shang\\
  \texttt{Chenming.Shang@dartmouth.edu}
}
\parbox[t]{0.5\textwidth}{
  \centering
  Yujin Tang\\
  \texttt{Yujin.Tang.GR@dartmouth.edu}
}\\
\AND
\parbox[t]{0.5\textwidth}{
  \centering
  Jun Jie Ou Yang\\
  \texttt{Jun.Jie.Ou.Yang.GR@dartmouth.edu}
}
\parbox[t]{0.5\textwidth}{
  \centering
  Ruize Xu\\
  \texttt{Ruize.Xu.GR@dartmouth.edu}
}\\
\AND
\parbox[t]{0.5\textwidth}{
  \centering
  Adam Breuer\textsuperscript{$\dagger$}\\
  \texttt{Adam.Breuer@dartmouth.edu}
}
\parbox[t]{0.5\textwidth}{
  \centering
  Nikhil Singh\textsuperscript{$\dagger$}\\
  \texttt{Nikhil.U.Singh@dartmouth.edu}
}\\
\\
Department of Computer Science, \textls[36]{Dartmouth College}\\
[8pt]\href{https://BreuerLabs.github.io/Platonic_Representation_Displacement/}{\faGlobe\ Project Homepage}
}

\begin{document}

\maketitle

\renewcommand{\thefootnote}{$\dagger$}
\footnotetext{Co-principal investigators.}

% !TEX root = ../neurips_2026.tex
\begin{abstract}
The Platonic Representation Hypothesis (PRH) claims that independently trained models converge on a shared statistical model of reality, yet recent work finds that the evidence weakens under some conditions.
In this paper, we measure what models share through the directions (displacement vectors) between samples. Under a single orthogonal alignment limited to just rotation and reflection, these displacement vectors are substantially preserved across $44$ independently trained vision and language encoders spanning modalities and asymmetric capability pairs, consistent with the PRH. Pointwise similarity, however, is lower. Representations split into a \emph{shared semantic component} that is linearly aligned across models, and a \emph{private component} that is not, which we interpret as having model-specific capabilities. We trace this geometry to concept-level structure: within a model, parent concepts are orthogonal to their child variation vectors; across models, concept displacements are parallel. We predict (and experiments confirm) that fine-tuning preserves Pointwise similarity but reduces displacement similarity, and relational distillation does the opposite. Because semantics align linearly but capabilities do not, we propose importing capabilities from one model to another using a single cached forward pass through the source. We call this \emph{Shadow Casting}. As a proof of concept, our \methodname instantiation outperforms strong fine-tuned baselines at far less compute. A cache can be released alongside open model weights, letting one model's capabilities be downloaded and imported into other models with frozen backbones.

% \renewcommand{\thefootnote}{*}
% \footnote{Project page: \url{https://helloscm.github.io/Platonic_Representation_Displacement/}.}

\end{abstract}
% !TEX root = ../neurips_2026.tex
\section{Introduction}
\label{sec:intro}
Two maps of the same city need not agree on coordinates to give you similar directions. One map may be cropped or drawn with a different convention for due north, so any given landmark may appear in a different absolute location on each page. Yet if they are good maps, then each will send a traveler off in the same direction to get from one landmark to another. That is, the displacement vectors between landmarks are preserved, even when their absolute positions on the page are not.

\textbf{The Platonic Representation Hypothesis and its discontents.}
Neural networks enjoy an analogous freedom: each learns its own coordinate system, so the absolute representation of a sample in one model need not match that of another. Indeed, measured directly, representations rarely simply align~\citep{sucholutsky2023getting}. The Platonic Representation Hypothesis (PRH) famously argues that despite this freedom, independently trained models converge toward a shared statistical model of reality~\citep{huh2024position}. If true, the implications are profound: it would mean that the choice of architecture, training objective, and even data modality is secondary to scale itself, so sufficiently capable models, regardless of how they are trained, all recover the same underlying structure. Yet very recent work challenges this core claim, finding that its supporting evidence degrades under more demanding evaluation and with larger-scale models~\citep{koepke2026back,groger2026revisiting}.

\textbf{Both sides define ``shared reality'' in terms of similar sample proximities across models.}
PRH uses pointwise neighborhoods. In this view, a shared reality exists when the samples nearest to a given point in one model's embedding space are also nearest to that point in another model's space. Given $n$ shared samples encoded by two models into representations $\boldsymbol{a}_i$ and $\boldsymbol{b}_i$, PRH measures:
\begin{equation}
    \text{mutual kNN}(i) = \frac{1}{k}\left|\mathcal{N}_k^{a}(i) \cap \mathcal{N}_k^{b}(i)\right|,
\end{equation}
where $\mathcal{N}_k^{a}(i)$ returns the $k$ samples closest to $i$ under model $a$'s inner product, and likewise for $b$. Under this definition, PRH finds increasing alignment: stronger models share more neighbors, and this sharing extends across modalities as models scale. Yet Huh et al.\ themselves note that CKA, which correlates the full pairwise similarity matrices $K_{ij} = \boldsymbol{a}_i^\top\boldsymbol{a}_j$ and $L_{ij} = \boldsymbol{b}_i^\top\boldsymbol{b}_j$, revealed only a very weak trend, so the convergence signal exists only in local neighborhoods, not in global structure. \citet{koepke2026back}'s test of this definition shows substantial degradation at scale, concluding that there exists no shared statistical model of reality on which learned representations converge.

% \textbf{Both sides measure where points sit, but neither measures where they point.}
The PRH evidence says models agree on which samples are close; the counter-evidence says they do not. But both define ``shared reality'' in terms of sample proximity, e.g., which points neighbor which others. Yet neither asks whether the vector between two samples points in the same direction across models. Returning to our analogy: would a traveler setting off from one sample toward another in Model~A head in the same direction as a traveler making this journey in Model~B?

In this paper, we show that independently trained models \emph{do} converge on a shared model of reality, which we identify using displacement directions rather than point proximities. Formally, we consider kernels defined on displacement vectors, which measure inter-model directional agreement:
\begin{equation}
    \cos(\boldsymbol{a}_i - \boldsymbol{a}_j,\;\boldsymbol{b}_i - \boldsymbol{b}_j).
\end{equation}
This quantity asks: after alignment, does the displacement from sample $i$ to sample $j$ in one model point in the same direction as the displacement between the same samples in another? 

\textbf{Main contributions.}
In this paper, we show that independently trained models share not the location of samples in representation space but \textit{the directions between them}. Specifically, we contribute:

\begin{enumerate}
    \item \textbf{Empirical evidence} that displacement geometry is preserved across $44$ vision and language encoders under trivial orthogonal alignment, giving a directional convergence criterion which complements proximity-based measures;
    \item A \textbf{structural account} of \textit{why} this geometry might emerge: within a model, hierarchical concepts exhibit an approximately orthogonal geometry~\citep[cf.][]{park2024geometry}, with parents (e.g., \emph{mammal}) orthogonal to their child displacements (e.g., $\boldsymbol{v}_{\text{dog}} - \boldsymbol{v}_{\text{mammal}}$); across models those displacements are parallel, in fact more parallel than sample-level displacements. We propose sample parallelism follows because samples are composed of concept increments;
    \item \textbf{Predictions, which experiments confirm:} fine-tuning preserves Pointwise similarity while reducing displacement similarity, and relational distillation~\citep[cf.][]{tung2019similarity} does the opposite. Thus, Pointwise and Cross-Model similarities respond differently to the choice of training objective;
    \item \textbf{Shadow Casting}: a capability-transfer method that exploits this geometry. Representations decompose into a shared semantic component that aligns linearly across models and a private capability component that does not, so capabilities can be imported from a source model via a single cached forward pass. We instantiate this in \textbf{\methodname}, which improves over the corresponding CLIP backbone at orders of magnitude less compute than full fine-tuning.
\end{enumerate}

% !TEX root = ../neurips_2026.tex
\section{Related Work}
\label{sec:related}

\paragraph{Representational alignment and the Platonic Representation Hypothesis.}

Representational alignment, i.e. determining when two systems have learned similar representations despite architectural or other differences, has a long history in cognitive science, neuroscience, and machine learning~\citep{edelman1998representation,kriegeskorte2008representational,muttenthaler2025aligning}. For a fuller picture of the field, we refer the reader to~\cite{sucholutsky2023getting}. This idea has recently been repopularized through the Platonic Representation Hypothesis (PRH), which argues that independently trained large neural networks may converge toward a shared statistical model of reality~\citep{huh2024position}. Subsequent work has characterized when alignment emerges and how it relates to task performance~\citep{tjandrasuwita2025understanding}, compared visual and linguistic concept representations through retrieval after Procrustes alignment~\citep{li2024vision}, extended alignment analysis to weight subspaces~\citep{kaushik2025universal}, and used it to inform method design~\citep{wang2025words,gupta2025better}. Theoretical work has also investigated PRH~\citep{lobashev2025information}, with recent results establishing sufficient conditions for cross-modal alignment in JEPAs~\citep{balestriero2026proving}. Other studies, however, highlight the sensitivity of measured cross-modal alignment to evaluation scale and non-bijective pairings~\citep{koepke2026back}, metric calibration~\citep{groger2026revisiting}, and stimulus selection~\citep{hosseini2026modulating}. Our work shifts the invariant under study from sample positions or similarity matrices to displacement directions, recovering a different view of shared reality.

\paragraph{Geometry of concept representations.}

The complementary question is how representations are arranged \textit{within} a single model~\citep{mikolov2013linguistic,nanda2023emergent}. The linear representation hypothesis~\citep{park2023linear,elhage2022toy} holds that semantic features are encoded as linear directions. \citet{park2024geometry} extend this, showing that simple categorical concepts are represented as simplices and that hierarchically related concepts are orthogonal under certain measures. Structural findings have also appeared at the parameter level~\citep{kaushik2025universal}, suggesting that geometric regularities may reflect something common to a broader class of models. While \textit{alignment} work characterizes when entire representation spaces agree and the concept-geometry literature characterizes how such spaces are arranged inside models, our work seeks to shed further light on \textit{cross-model} concept geometry and its potential for analyzing convergence claims.

\paragraph{Transferring knowledge between models.}
Other work uses representational alignment as an engineering resource, a mechanism for \textit{transfer}~\citep{moschella2022relative,maiorca2023latent,norelli2023asif}. Classical knowledge distillation transfers behavior from a teacher to a student through output matching~\citep{hinton2015distilling}. Later, it was shown that preserving similarity structure in hidden representations can transfer relational knowledge~\citep{tung2019similarity}. More recent work studies model-agnostic knowledge transfer, showing that arbitrary pretrained models can contain complementary knowledge that can sometimes be transferred even across performance gaps~\citep{roth2023fantastic}. 
%Often, such methods implicitly presuppose that meaningful structure is shared and complementary across models. Our analysis of displacement directions makes this shared structure geometrically explicit and helps explain why such transfer techniques work. We then use these principles to design and validate a new transfer method.
Such methods often implicitly presuppose shared structure and complementary knowledge across models. Our analysis of displacement directions makes the shared geometry explicit, helps explain why such transfer techniques work, and guides the design of a new transfer method.
% !TEX root = ../neurips_2026.tex
\section{Measuring Shared Reality Directionally}
\label{sec:prelim}

We now show how to measure shared reality directionally. Models produce representations of different dimensionalities, scales, and coordinate systems, so any cross-model comparison requires preprocessing and alignment. To the extent that shared reality exists, it should not require sophisticated tools to recover. We therefore restrict ourselves to the most minimal standard procedures at each step: PCA to a common dimension, whitening to remove anisotropy, and a single rigid Procrustes rotation/reflection. Additionally, existing cross-model similarity measures evaluate kernels on inputs, not on \textit{input pairs}, and so we discuss our procedure for doing the latter below. Full preprocessing details and robustness checks are in App.~\ref{app:prelim}.

\textbf{Setup.}
A representation is a map $f_m : \mathcal{X} \to \mathbb{R}^{d_m}$ from inputs to features, and $\boldsymbol{m}_i \coloneqq f_m(i)$ is the representation of input $i$ in model $m$. Our central object is the \textit{displacement vector} between two inputs, $\boldsymbol{d}_m(i,j) \coloneqq f_m(i) - f_m(j) \in \mathbb{R}^{d_m}$. Existing similarity measures reduce to comparisons of input-level kernels: CKA compares each model's within-model similarity structure, and mutual-$k$NN compares neighborhoods. We evaluate kernels on input \textit{pairs}, asking whether the \textit{direction} of $\boldsymbol{d}_m(i,j)$ is preserved across $m$.

\textbf{Alignment.} For each pair of models $a, b$ we fit a single orthogonal Procrustes rotation $R_{ab}$ on a held-out subset of inputs and evaluate metrics on the disjoint complement. Let $\widehat{X}_a,\widehat{X}_b\in\mathbb{R}^{n_{\mathrm{fit}}\times d}$ contain the PCA-reduced, whitened representations of the matched alignment inputs. We solve
\begin{equation}
    R_{ab}=\arg\min_{R^\top R=I}\left\|\widehat{X}_aR-\widehat{X}_b\right\|_F,
    \label{eq:opa-main}
\end{equation}
with solution $R_{ab}=UV^\top$, where $U\Sigma V^\top=\widehat{X}_a^\top\widehat{X}_b$. The alignment is rigid (rotation and reflection only, with no scaling or additional learned mapping) in the PCA-reduced, whitened space, and the same $R_{ab}$ is used for all three metrics below, so no metric is individually optimized. After alignment, each model's representations are shifted by its mean, mapped through its own PCA and whitening (shift is the same for every input, so it cancels in Cross-Model but factors into Pointwise and Cross-Object). We write $\boldsymbol{a}_i, \boldsymbol{b}_i$ for the post-alignment representations of input $i$ in models $a, b$.

\begin{wrapfigure}{r}{0.35\linewidth}
    \centering
    
    \includegraphics[width=\linewidth]{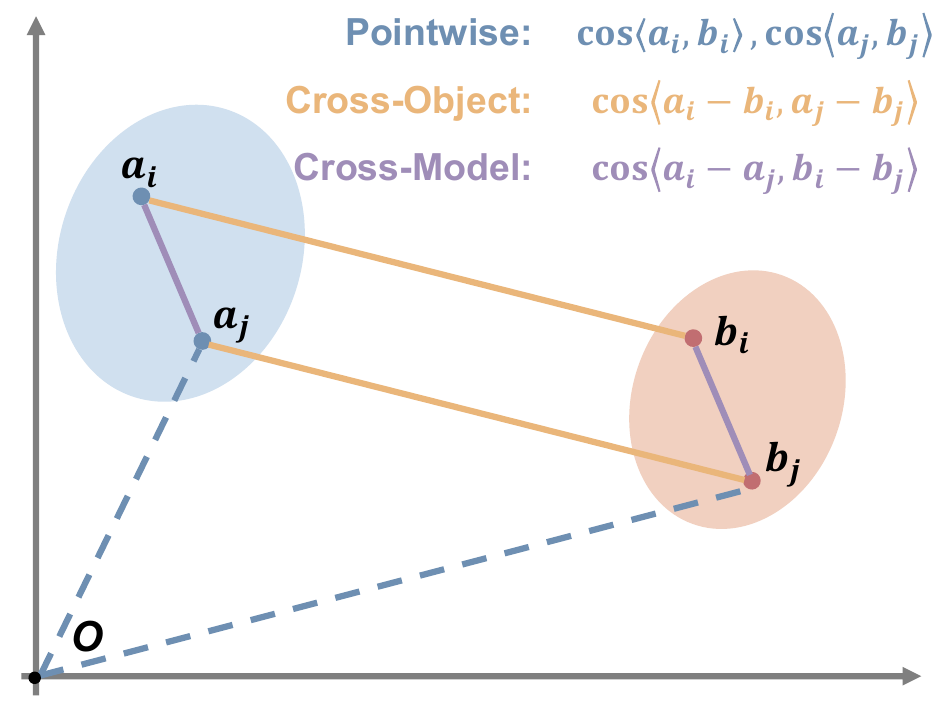}
    
    \caption{We compute three cosine measures of cross-model agreement on inputs $i, j$ with aligned representations $\boldsymbol{a}_i, \boldsymbol{a}_j$ (model $a$) and $\boldsymbol{b}_i, \boldsymbol{b}_j$ (model $b$). \textbf{Pointwise} (blue): $\cos(\boldsymbol{a}_i, \boldsymbol{b}_i)$. \textbf{Cross-Object} (orange): $\cos(\boldsymbol{a}_i - \boldsymbol{b}_i,\; \boldsymbol{a}_j - \boldsymbol{b}_j)$, a cross-model offset control. \textbf{Cross-Model} (purple): $\cos(\boldsymbol{a}_i - \boldsymbol{a}_j,\; \boldsymbol{b}_i - \boldsymbol{b}_j)$, the directional convergence quantity described in \S\ref{sec:intro}.}
    \label{fig:metric-schematic}
    
\end{wrapfigure}

\textbf{Metrics.} For inputs $i, j$ we compute three cosine quantities:

\hspace*{1em}$\bullet$~\textbf{Pointwise}: $\cos(\boldsymbol{a}_i,\, \boldsymbol{b}_i)$ measures angular agreement between the two representations of the same input (including each model's restored mean).

\hspace*{1em}$\bullet$~\textbf{Cross-Object}: $\cos(\boldsymbol{a}_i - \boldsymbol{b}_i,\;\, \boldsymbol{a}_j - \boldsymbol{b}_j)$ asks whether the offset between the two models on input $i$ matches their offset on a different input $j$. This is large whenever the two models' restored means differ, regardless of training.

\hspace*{1em}$\bullet$~\textbf{Cross-Model}: $\cos(\boldsymbol{a}_i - \boldsymbol{a}_j,\;\, \boldsymbol{b}_i - \boldsymbol{b}_j)$ asks whether moving from input $i$ to input $j$ traces the same direction in both models, i.e., whether the models agree on \emph{where things point}. This is the directional convergence quantity introduced in §\ref{sec:intro}.

\textbf{Cross-model consensus decomposition.}
To recover the shared semantic and private capability subspaces discussed in \S\ref{sec:intro}, we decompose each aligned representation $Z_m$ into a shared and a private subspace via the eigendecomposition of a cross-model agreement matrix $S_{\mathrm{cross}}$ (App.~\ref{app:prelim}, eq.~\ref{eq:scross}). Here, capability refers to task-relevant functional specializations, such as CLIP's sensitivity to language-described semantics and DINO's fine-grained visual understanding. For decomposition, we jointly align the $M$ models using generalized Procrustes analysis (GPA), obtaining $Z_m=\widehat{X}_mR_m\in\mathbb{R}^{n\times d}$ for $n$ shared inputs. We compute
\begin{equation}
    S_{\mathrm{cross}}=\frac{1}{M(M-1)}\sum_{a\ne b}\frac{Z_a^\top Z_b+Z_b^\top Z_a}{2(n-1)}.
    \label{eq:scross-main}
\end{equation}
$S_{\mathrm{cross}}$ scores directions by cross-model consistency: top-$r$ eigenvectors span the shared subspace, bottom-$r$ the private one. Writing the corresponding orthonormal basis matrices as $P_r^{\mathrm{com}},P_r^{\mathrm{priv}}\in\mathbb{R}^{d\times r}$, we obtain the label-free decomposition
\begin{equation}
    Z_m^{\mathrm{com}}(r)=Z_mP_r^{\mathrm{com}},
    \qquad
    Z_m^{\mathrm{priv}}(r)=Z_mP_r^{\mathrm{priv}}.
    \label{eq:decomp-main}
\end{equation}
% !TEX root = ../neurips_2026.tex
\section{Experiments: Geometry of Representation Displacements}
\label{sec:experiments}
Our goal in this section is to characterize the displacement geometry that independently trained models share. Specifically, three empirical claims follow from our framework and reframe the convergence debate. First, displacement geometry is shared with high fidelity across independently trained models, most strikingly at the concept level. Second, displacement similarity and Pointwise similarity respond differently to the training objective. Third, representations admit an explicit decomposition into a shared semantic component that aligns linearly across models and a private capability component that does not. Together these results show that displacement directions agree across models. \S\ref{sec:concept_main}--\ref{ssec:samplelevel} ask what geometric structure independently trained models share, especially at the concept level. \S\ref{sec:displacement}--\ref{sec:decomp} then localize that structure across model depth, training objectives, and shared versus model-specific subspaces. Finally, Section 5 tests whether this shared structure can guide the transfer of source-specific information through our Shadow Casting proof of concept.

All results use \emph{rigid alignment only} after per-model PCA/whitening: orthogonal Procrustes transformations, without scaling, learned mappings, or nonlinearity. Under more permissive alignment, high similarity would be unsurprising; our claim is that it is strong enough that no such machinery is required to observe it.

\textbf{Setup.} We evaluate 44 pretrained models: 25 vision transformers spanning three sizes (base, large, huge) and six pretraining paradigms (supervised, CLIP, DINO, MAE, SigLIP, I-JEPA), plus 19 LLMs (Llama, Gemma, Mistral, OLMo, Bloomz). We denote vision--vision, language--language, and vision--language model pairs by V-V, L-L, and V-L, respectively. We extract hidden states after each complete transformer block, including both attention and MLP updates. For LLMs, we mean-pool over valid token positions; for vision transformers, we use the model's standard pooling rule, either CLS or global average pooling. Sample-level analysis uses Flickr30k, whose paired image-caption pairs allow shared sample indexing across vision and language models. Concept-level analysis uses the intersection of ImageNet-22K and CommonWords79k~\citep{li2024vision}: each ImageNet-22K class is a concept, represented by the centroid over its members; hierarchical (parent, child, sibling) relationships are extracted from WordNet. All inter-model comparisons use a single orthogonal Procrustes rotation per model pair, fit on a held-out split and evaluated on the disjoint complement; the same rotation is used for all three metrics on a given model pair, so no metric is individually optimized. We compare against randomly initialized models with different seeds as a null for architectural bias. Apps.~\ref{app:addlexperiments}--\ref{app:experiments} give full details, including robustness across  modalities (\ref{app:heatmapsection}), concepts (\ref{app:concept-orthogonality}), and scale, including asymmetric model capability pairs (\ref{app:asym}).

\textbf{Replication code.} Full replication code for all experiments is available at
\url{https://github.com/BreuerLabs/Platonic_Representation_Displacement}.
% !TEX root = ../neurips_2026.tex
\begin{figure}[t]
    \centering
    \begin{subfigure}[b]{0.32\linewidth}
        \centering
        \includegraphics[width=\linewidth]{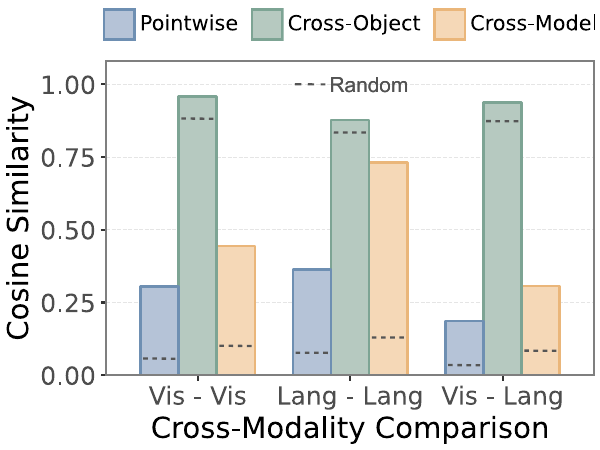}
        \caption{Sample-level}
        \label{fig:bar-sample}
    \end{subfigure}
    \hfill
    \begin{subfigure}[b]{0.32\linewidth}
        \centering
        \includegraphics[width=\linewidth]{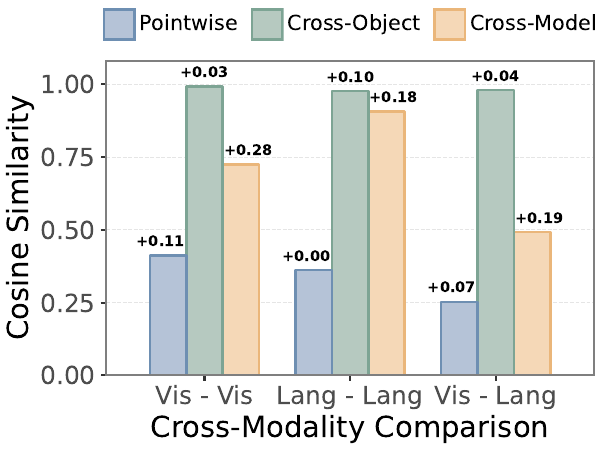}
        \caption{Concept-level}
        \label{fig:bar-concept}
    \end{subfigure}
    \hfill
    \begin{subfigure}[b]{0.32\linewidth}
        \centering
        \includegraphics[width=\linewidth]{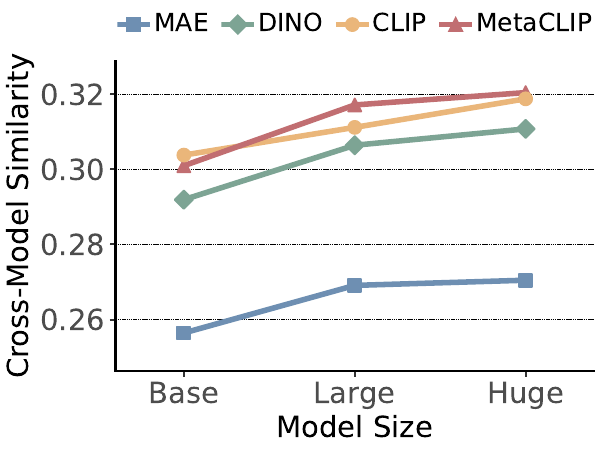}
        \caption{Effect of model scale}
        \label{fig:bar-scale}
    \end{subfigure}
    \caption{Cross-Model parallelism analysis.}
    
    \label{fig:bar-triplet-sample-concept-scale}
\end{figure}
\subsection{Concept-level displacement geometry is shared across independently trained models}
\label{sec:concept_main}

Under rigid alignment alone, displacement vectors between concepts are preserved with high fidelity across independently trained models. Fig.~\ref{fig:bar-concept} shows concept-level Cross-Model similarity of $0.73$ (V-V), \ $0.91$ (L-L), \ $0.49$ (V-L).  The mean cosine between $(\boldsymbol{v}_{\text{dog}} - \boldsymbol{v}_{\text{cat}})$ in, e.g., Llama and the same displacement in Gemma is 0.91---substantially aligned, after only a rigid rotation. By contrast, Pointwise similarity at the concept level is lower at $0.41$ (V-V), \ $0.36$ (L-L), \ $0.25$ (V-L). Paired Wilcoxon signed-rank tests confirm that Cross-Model exceeds Pointwise by a median of $0.41 \pm 0.011$ (V-V), $0.47 \pm 0.011$ (L-L), and $0.24 \pm 0.003$ (V-L). All are significant at $p < 0.0001$ (see App. \ref{app:statistics}).

\textbf{Concept-level parallelism exceeds sample-level parallelism.} Fig.~\ref{fig:bar-concept} reports the increase from sample-level to concept-level Cross-Model similarity: $+$0.28 (V-V), \ $+$0.18 (L-L), \ $+$0.19 (V-L). This is consistent with an inherited geometry: if samples are weighted combinations of concepts and concept-level structure is shared, then sample-level structure is a noisier reflection of it.

\textbf{Cross-Object as a cross-model offset control.} \emph{Cross-Object} similarity is high across all conditions ($0.87$-$0.96$), including for randomly initialized models (Fig.~\ref{fig:bar-sample}). It asks whether the cross-model offsets for two inputs point in the same direction, which can occur even in untrained networks when models' means differ. \emph{Cross-Model} isolates \emph{learned} convergence: it is near zero for random networks (gray dashed lines in Fig.~\ref{fig:bar-sample}) and substantial for trained ones, indicating that displacement convergence reflects what models learn rather than what their architecture imposes.

Overall, Cross-Model provides a directional criterion that complements the proximity-based measures used by PRH and its critics.

% \textbf{Reconciling directions vs. positions.} The concept-level result clarifies the distinction between two notions of agreement. Both measured \emph{positions}, i.e., which samples land in similar neighborhoods, and reached opposite conclusions about whether convergence holds. Positions differ across models (low Pointwise similarity), while directions between concepts agree (high Cross-Model similarity, consistent with the PRH intuition~\citep{huh2024position}). Cross-Model thus provides a directional convergence criterion that complements proximity-based measures prominently used by PRH and its critics.

\subsection{Concept geometry explains how cross-model parallelism is inherited by samples}
\label{sec:orthogonality}
\begin{wrapfigure}{r}{0.385\linewidth}
    \centering
    \vspace{-\baselineskip}
    \includegraphics[width=\linewidth]{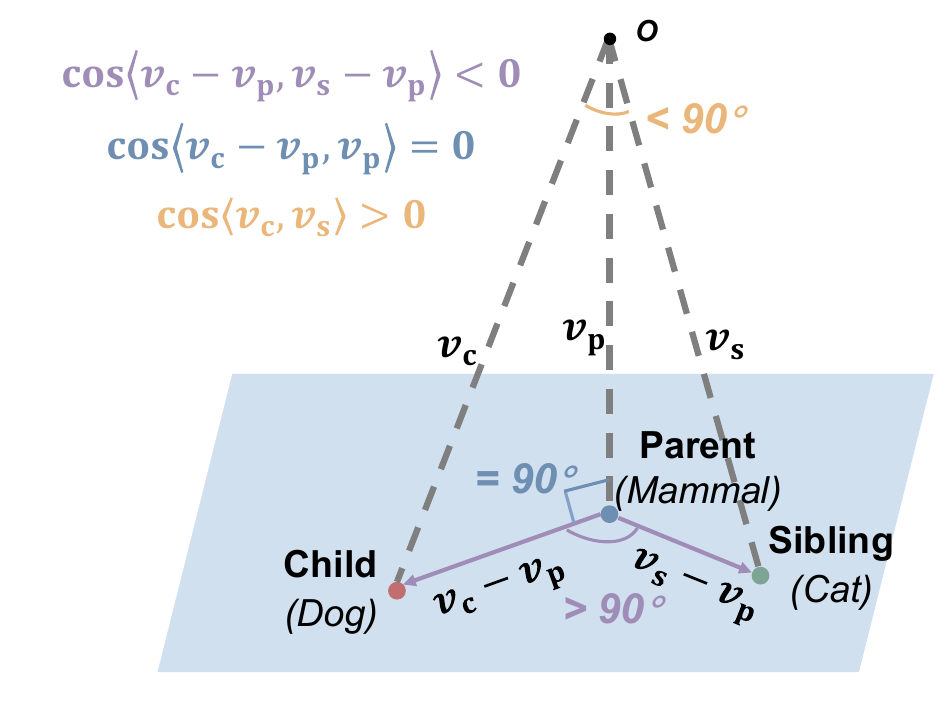}
    \caption{Hierarchical concept geometry: child variations are orthogonal to the parent concept.}
    \label{fig:ortho-schematic}

    \vspace{0.5\baselineskip}
    \includegraphics[width=\linewidth]{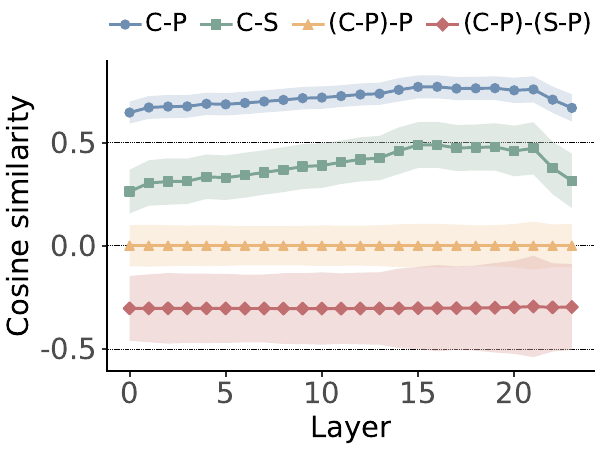}
    \caption{Parent--child and sibling relationships persist across model layers.}
    \label{fig:ortho-layer}
    \vspace{-1.5\baselineskip}
\end{wrapfigure}
We now investigate how concept-level geometry gives rise to sample-level displacement parallelism. We trace the result to a more elementary geometric property of how concepts are organized within each model: hierarchical concepts exhibit an approximately orthogonal geometry, where parent concepts and the variation vectors that distinguish their children occupy orthogonal subspaces.

\textbf{Three regularities of within-model concept geometry.} We observe three intra-model geometric regularities (Figs.~\ref{fig:ortho-schematic} and~\ref{fig:ortho-layer}). First, parent and child concepts (e.g., \emph{mammal}, \emph{dog}) have positive cosine similarity ($\cos(\boldsymbol{v}_{\text{child}}, \boldsymbol{v}_{\text{parent}}) \approx 0.72$, mean across layers and WordNet pairs): children retain their parent's representational support. Second, the variation vector from parent to child is orthogonal to the parent: $\cos(\boldsymbol{v}_{\text{child}} - \boldsymbol{v}_{\text{parent}}, \boldsymbol{v}_{\text{parent}}) \approx 0$. The features that distinguish the child develop in an orthogonal subspace, leaving the parent undisturbed. Third, within this orthogonal subspace siblings anti-align with each other: $\cos(\boldsymbol{v}_{\text{child}} - \boldsymbol{v}_{\text{parent}}, \boldsymbol{v}_{\text{sibling}} - \boldsymbol{v}_{\text{parent}}) \approx -0.30$, achieving discriminability.

\textbf{Layer-wise persistence.} These regularities hold at every layer (Fig.~\ref{fig:ortho-layer}): parent-child positive (C-P, blue), child-sibling positive but smaller (C-S, green), child-parent vs.\ parent near zero (yellow), child-parent vs.\ sibling-parent negative (red). Concept orthogonality is therefore a structural feature of the representation, not a property of any particular layer.

\textbf{Samples inherit cross-model parallelism from concepts.} With the cross-model concept parallelism of §\ref{sec:concept_main}, within-model orthogonality explains the inheritance from concept-level to sample-level structure. If samples can be modeled as weighted combinations of concepts, and (i) within-model concept geometry is approximately orthogonal, (ii) cross-model concept displacements are parallel, and (iii) models weight concepts similarly for each sample, then sample-level displacements should be approximately parallel as well. The $0.18$--$0.28$ gap between concept-level and sample-level Cross-Model similarity (Fig.~\ref{fig:bar-concept}) is the empirical signature: concept-level dominates, and sample-level is a noisier reflection. App. \ref{app:failure} further shows that, in simulation, degrading concept-level geometry lowers sample-level parallelism.

\subsection{Sample-level results corroborate the concept-level finding}
\label{ssec:samplelevel}
Sample-level Cross-Model similarity exceeds Pointwise across V-V, L-L, and V-L pairs, but more weakly than at the concept level (Fig.~\ref{fig:bar-sample}): Pointwise averages $0.31$, \ $0.36$, \ $0.19$; Cross-Model averages $0.45$, \ $0.73$, \ $0.31$. The L-L gap is the largest (roughly 2$\times$); the V-L gap is the smallest. 
Paired Wilcoxon signed-rank tests confirm that Cross-Model exceeds Pointwise by a median of $0.23 \pm 0.008$ (V-V), $0.39 \pm 0.007$ (L-L), and $0.15 \pm 0.002$ (V-L). All are significant at $p < 0.0001$ (App. \ref{app:statistics}).
Cross-Model similarity also increases with model size (Fig.~\ref{fig:bar-scale}) and depends on pretraining objective: CLIP and DINO yield higher Cross-Model similarity than MAE, consistent with their emphasis on global semantic structure rather than pixel-level reconstruction. Holding pretraining objective fixed, dataset choice has little effect.
% !TEX root = ../neurips_2026.tex
\begin{figure}[t]
    \centering
    \begin{minipage}{\textwidth}
    \centering
    \begin{subfigure}[b]{0.32\linewidth}
        \centering
        \includegraphics[width=\linewidth]{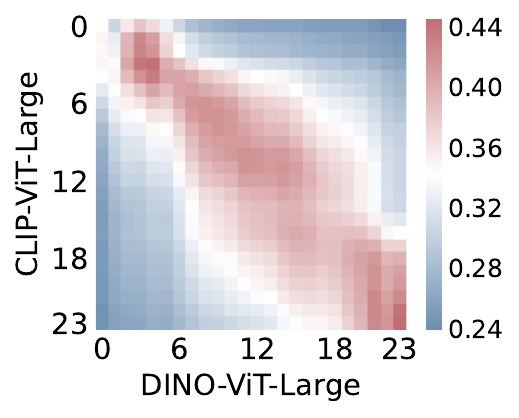}
        \caption{V-V}
        \label{fig:heatmap-vv}
    \end{subfigure}
    \hfill
    \begin{subfigure}[b]{0.32\linewidth}
        \centering
        \includegraphics[width=\linewidth]{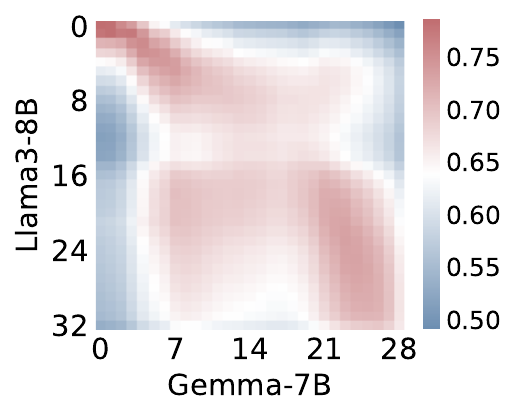}
        \caption{L-L}
        \label{fig:heatmap-ll}
    \end{subfigure}
    \hfill
    \begin{subfigure}[b]{0.32\linewidth}
        \centering
        \includegraphics[width=\linewidth]{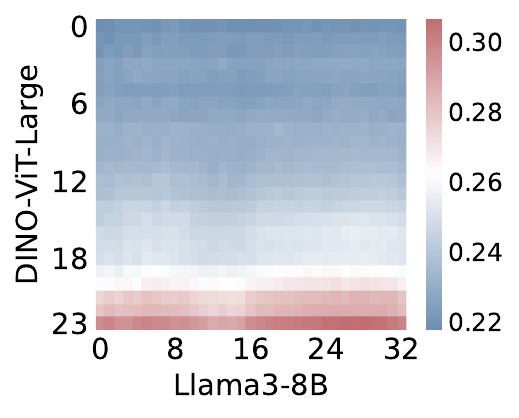}
        \caption{V-L}
        \label{fig:heatmap-vl}
    \end{subfigure}
    \end{minipage}
    \caption{Cross-Model similarity comparison across all layers of different models.}
    \label{fig:triple-heatmap}
\end{figure}

\textbf{Robustness to asymmetric capability and to CLIP.} Cross-Model similarity remains high across model pairs of very different scale and capability, e.g., CLIP-ViT-Huge vs.\ MAE-ViT-Base: 0.38; Llama3-8B vs.\ Bloomz-560M: 0.61. Displacement parallelism is therefore not a consequence of comparing models of similar strength. To rule out that V-L convergence is a CLIP artifact, we compare DINO directly against Llama: Cross-Model similarity is 0.306 (DINO-Large vs.\ Llama3-8B), almost identical to CLIP-Large vs.\ Llama3-8B (0.311). V-L convergence is therefore not driven by CLIP's vision-language objective. App.~\ref{app:asym} reports full results.

\subsection{Displacement convergence is concentrated where semantic content lives}
\label{sec:displacement}
Figs.~\ref{fig:heatmap-vv}, \ref{fig:heatmap-ll}, \ref{fig:heatmap-vl} show Cross-Model similarity between every pair of layers across two models; the pattern differs systematically by modality. For V-V pairs (Fig.~\ref{fig:heatmap-vv}), high similarity concentrates along the diagonal: layer $\ell$ in Model~A shares displacement geometry with layer $\ell$ in Model~B. For L-L pairs (Fig.~\ref{fig:heatmap-ll}), similarity extends broadly across the layer grid, especially in middle and deep layers. For V-L pairs (Fig.~\ref{fig:heatmap-vl}), nearly every language-model layer aligns with deep vision-model layers.

The asymmetry is informative. In vision transformers, semantic structure emerges progressively from low-level visual features in early layers to abstract semantic features in deep layers \citep{caron2021emerging}. In language models, semantic structure carries through the depth from the first token-embedding layer onward. The cross-modal pattern, in which language layers align with deep vision layers but not shallow ones, is what shared \emph{semantic} structure predicts, consistent with §\ref{sec:decomp}.

\subsection{Training objectives affect Pointwise and Cross-Model similarity differently}
\label{sec:training_dynamics}
If Pointwise and Cross-Model similarity were two views of the same underlying object, no training procedure could move one without moving the other. We test this by tracking both quantities at every epoch, refitting preprocessing and alignment each time, under three training regimes that target position vs.\ relation: standard fine-tuning (or supervised training from scratch), feature distillation (FKD), and relational distillation (RKD).

\textbf{Predictions.} (i) Fine-tuning should preserve Pointwise (representations keep their original location) while reducing Cross-Model (relational structure is reorganized for the task). Prior studies relatedly find that fine-tuning retains aspects of pretrained structure while introducing task-dependent changes~\citep{zhou2022closer,merlin2023finetuning}. (ii) Feature distillation, which trains a student to match teacher representations directly, should increase Pointwise (positions are pulled toward the teacher) and consequently Cross-Model (relations follow). (iii) Relational distillation, which trains a student to match teacher inter-sample similarities, should increase Cross-Model (relations are directly optimized) without increasing Pointwise (positions are unconstrained).

\begin{figure}[t]
    \centering
    \begin{minipage}{\columnwidth}
    \centering
    \begin{subfigure}[b]{0.32\linewidth}
        \centering
        \includegraphics[width=\linewidth]{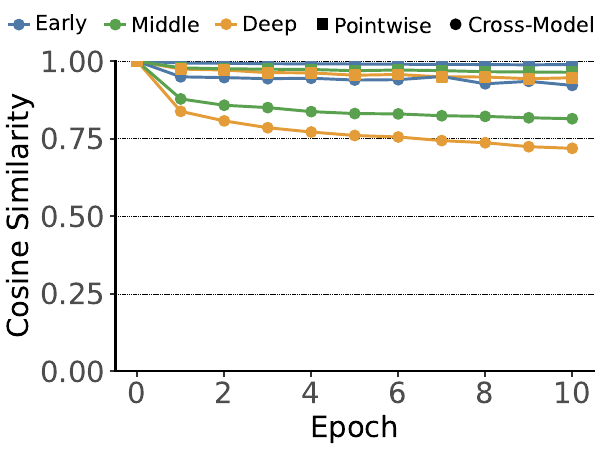}
        \caption{Finetune}
        \label{fig:dynamics-finetune}
    \end{subfigure}
    \hfill
    \begin{subfigure}[b]{0.32\linewidth}
        \centering
        \includegraphics[width=\linewidth]{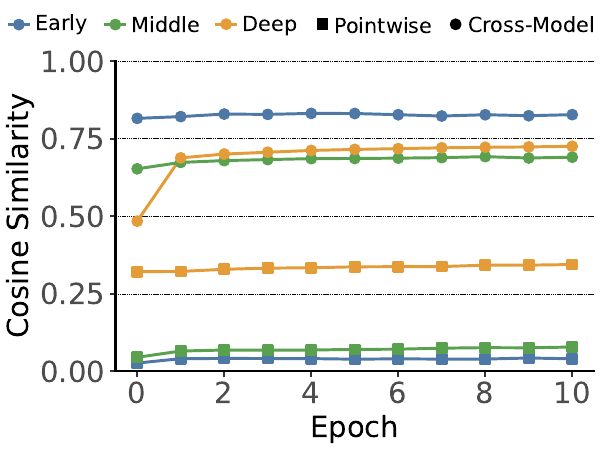}
        \caption{Relational distillation}
        \label{fig:dynamics-rkd}
    \end{subfigure}
    \hfill
    \begin{subfigure}[b]{0.32\linewidth}
        \centering
        \includegraphics[width=\linewidth]{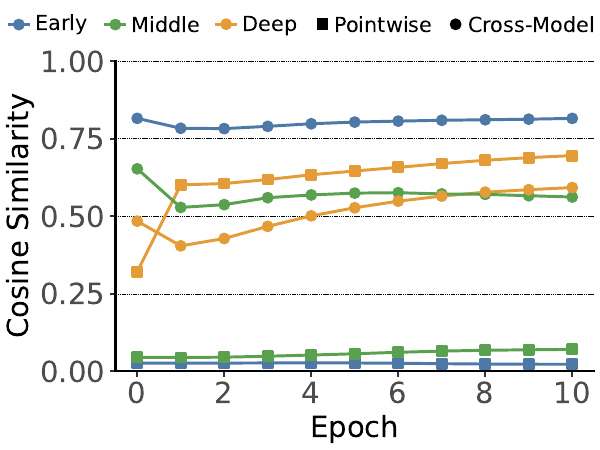}
        \caption{Feature distillation}
        \label{fig:dynamics-fkd}
    \end{subfigure}

    \begin{subfigure}[b]{0.32\linewidth}
        \centering
        \includegraphics[width=\linewidth]{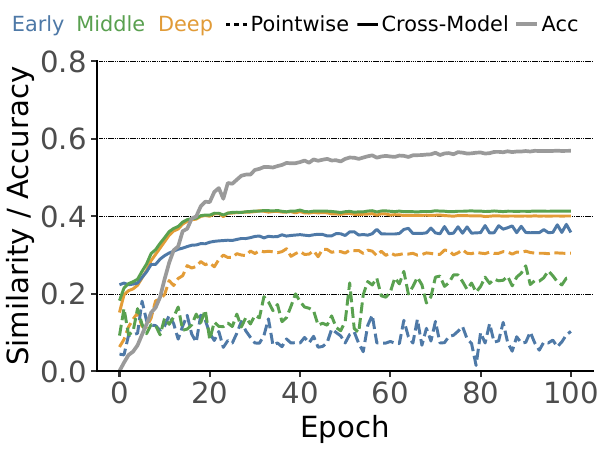}
        \caption{Scratch}
        \label{fig:scratch-raw}
    \end{subfigure}
    \hfill
    \begin{subfigure}[b]{0.32\linewidth}
        \centering
        \includegraphics[width=\linewidth]{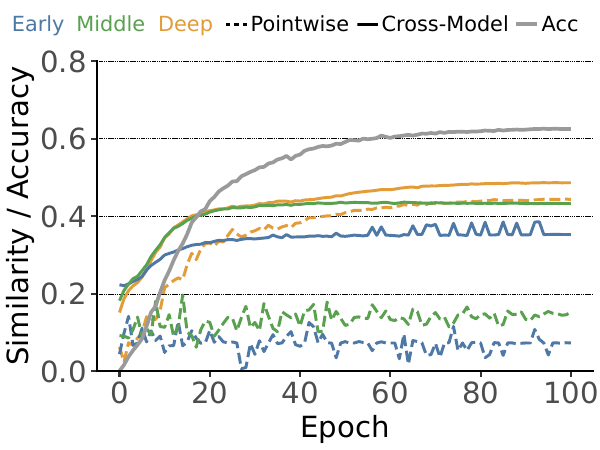}
        \caption{Scratch + relational distillation}
        \label{fig:scratch-rkd}
    \end{subfigure}
    \hfill
    \begin{subfigure}[b]{0.32\linewidth}
        \centering
        \includegraphics[width=\linewidth]{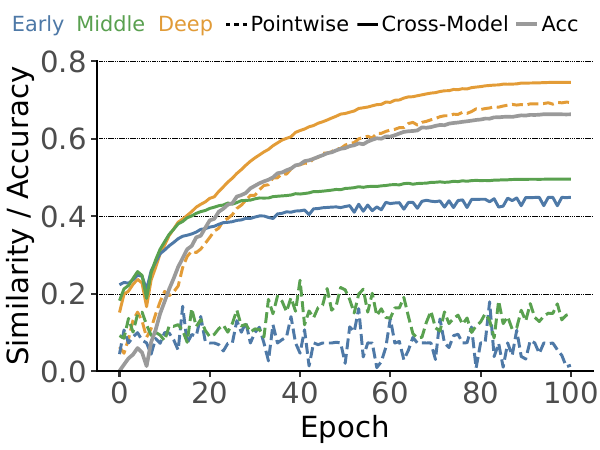}
        \caption{Scratch + feature distillation}
        \label{fig:scratch-fkd}
    \end{subfigure}
    \end{minipage}
    \caption{Evolution of Pointwise and Cross-Model similarity under different training objectives.}
    \label{fig:dynamicsGRIDplot}
\end{figure}

\textbf{Setup.} We test these predictions in two settings (Figs.~\ref{fig:dynamics-finetune}--\ref{fig:scratch-fkd}). \emph{Top row, from pretrained models, 10 epochs.} (a) Fine-tune: CLIP-ViT-B/16 fine-tuned on ImageNet-1K, evaluated against the original CLIP-ViT-B/16. (b) RKD: supervised ViT-B/16 student distilled against CLIP-ViT-B/16 teacher with distillation loss only, evaluated against the CLIP teacher. (c) FKD: same as (b) but feature distillation. \emph{Bottom row, from scratch, 100 epochs.} (d) Scratch: randomly-initialized ViT-S/16 trained on ImageNet-1K with classification loss. (e) Scratch+RKD: same as (d) plus relational distillation against supervised ViT-B/16. (f) Scratch+FKD: same as (d) plus feature distillation. RKD uses KL divergence on pairwise cosine-similarity distributions; FKD uses MSE on representations.

\textbf{Results.} Figs.~\ref{fig:dynamics-finetune}--\ref{fig:scratch-fkd} confirm each prediction in both rows. Top row (10 epochs from pretrained models): fine-tuning leaves Pointwise nearly unchanged ($<0.05$ over $10$ epochs) while dropping Cross-Model in the deep layers from $1.0$ to $0.72$; FKD raises both in deep layers (Pointwise $0.32 \to 0.70$, Cross-Model $0.48 \to 0.62$); RKD raises only Cross-Model ($0.48 \to 0.74$), leaving Pointwise flat ($0.32$--$0.35$). 
Bottom row (100 epochs from scratch): supervised training increases both similarities; RKD further improves Cross-Model alignment; FKD yields larger gains in both, with a particularly pronounced increase in Pointwise similarity.
The two similarities therefore need not change together; their trajectories depend on the training objective. Thus, Pointwise and Cross-Model similarity respond differently to the training objective, and are not reducible to one another.

\subsection{Representations decompose: linearly-shared semantics + nonlinearly-private capabilities}
\label{sec:decomp}

% !TEX root = ../neurips_2026.tex
% !TEX root = ../neurips_2026.tex
\begin{figure}[t]
    \centering
    \begin{minipage}{\textwidth}
    \centering
    \begin{subfigure}[b]{0.32\linewidth}
        \centering
        \includegraphics[width=\linewidth]{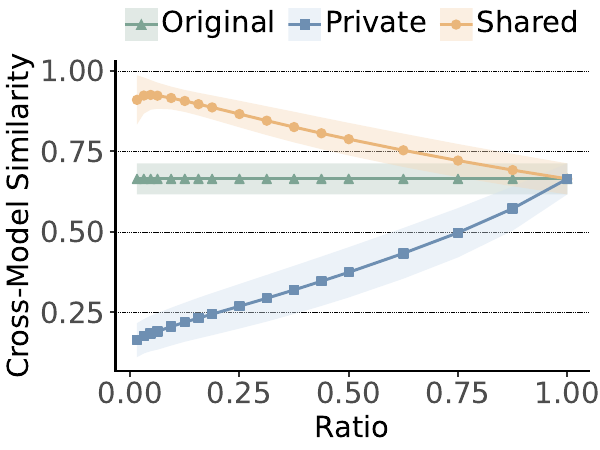}
        \caption{\mbox{Cross-Model similarity}}
        \label{fig:decomp-cross}
    \end{subfigure}
    \hfill
    \begin{subfigure}[b]{0.32\linewidth}
        \centering
        \includegraphics[width=\linewidth]{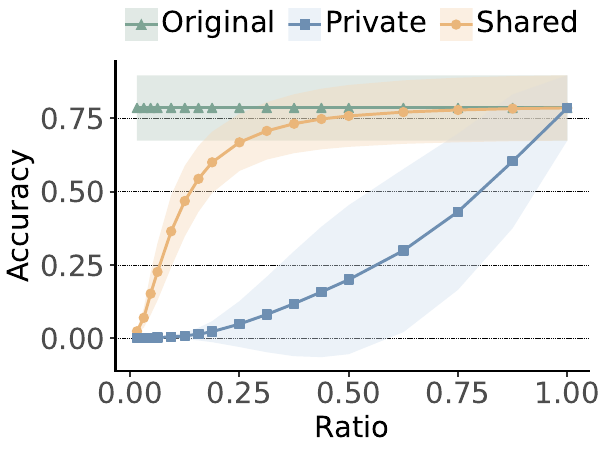}
        \caption{Linear probe accuracy}
        \label{fig:decomp-lp}
    \end{subfigure}
    \hfill
    \begin{subfigure}[b]{0.32\linewidth}
        \centering
        \includegraphics[width=\linewidth]{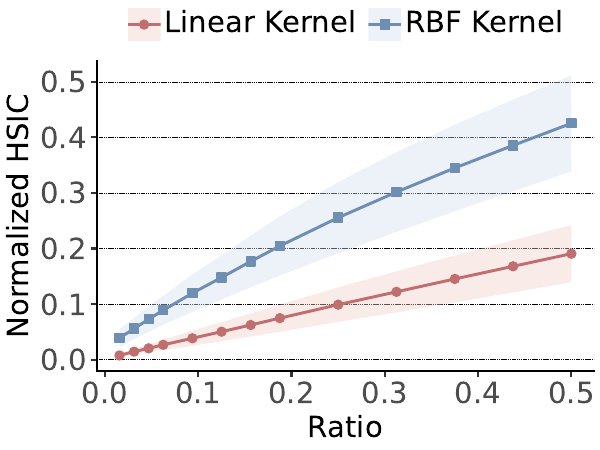}
        \caption{HSIC, linear vs.~RBF}
        \label{fig:decomp-hsic}
    \end{subfigure}
    \end{minipage}
    \caption{Metrics for the shared and private components after decomposition.}
\end{figure}

Under the cross-model consensus decomposition (§\ref{sec:prelim}), representations resolve into two cleanly separated components: a shared component that is linearly aligned across models and semantically rich, and a private component that is linearly opaque but statistically dependent on the shared one.
Figs.~\ref{fig:decomp-cross}, \ref{fig:decomp-lp}, and \ref{fig:decomp-hsic} report three properties of the decomposition. Fig.~\ref{fig:decomp-cross} shows Cross-Model similarity as the subspace dimension ratio $r/d$ varies from 0 to 1: at matched dimensions, the shared component has higher similarity than the private component, whose similarity increases with $r/d$. Fig.~\ref{fig:decomp-lp} shows linear probe accuracy on the ImageNet-1K validation set for each component: the shared subspace retains more linearly decodable semantic content at matched dimensions, while the private subspace has low accuracy at small $r/d$. For both metrics, the scores of both components approach those of the original representation as $r/d$ approaches 1, where both subspaces span the full space. Fig.~\ref{fig:decomp-hsic} shows normalized HSIC~\citep{gretton2005measuring} between shared and private components under linear and RBF kernels (see \hyperref[app:hsic]{App.~\ref*{app:prelim}} for definitions and computation details): linear dependence is low (due to whitening); RBF dependence is substantially higher (persistent nonlinear entanglement). The private component is therefore statistically dependent on the shared one.  %%
% !TEX root = ../neurips_2026.tex

\section{Shadow Casting: Capability Transfer via a Cached Forward Pass}
\label{sec:shadowclip}

Our experiments motivate a transfer principle: because the semantic component aligns linearly across models but the capability component does not, capabilities can be \emph{imported} from a source model by caching source representations in a single forward pass over the training set and training lightweight modules on the cached representations. We call this \emph{casting the shadow} of each training sample into the source's representation space, and the resulting transfer principle \textbf{Shadow Casting}. As a proof of concept, we instantiate Shadow Casting in \methodname{}.

% The decomposition also motivates the operator's form: \emph{linear} for capabilities carried by linearly-aligned semantic structure (as in language), \emph{nonlinear} for capabilities nonlinearly entangled with semantics (as in vision). The shadow is computed once and can be released alongside open weights, letting any number of downstream models import the source's capabilities without re-training the source model. Unlike adapters, LoRA, or model merging, Shadow Casting requires no further access to the source model during training once the representations are cached.

\textbf{\methodname.} \methodname{} is a vision-language model with a frozen CLIP backbone augmented by two lightweight residual modules (full architecture in App.~\ref{app:shadowclip}). On the vision side, a SwiGLU \citep{shazeer2020glu} residual takes the CLIP representation as a gating signal and the DINO representation as a value, injecting nonlinearly-entangled fine-grained visual structure that vanilla CLIP lacks. On the language side, since §\ref{sec:training_dynamics} showed that relational distillation from a pretrained model raises Cross-Model similarity while Pointwise stays flat, we attach a two-layer MLP residual after the CLIP text encoder and supervise it via relational distillation against an LLM teacher (LLM2Vec-Gen-Qwen3-4B), using KL divergence on the pairwise cosine-similarity distributions of (whitened) student and teacher representations. At inference, the LLM teacher is no longer needed, while the vision branch still uses DINO to encode unseen images. Both residuals are gated by a learnable vector initialized to zero, so \methodname{} is exactly CLIP at initialization.

% !TEX root = ../neurips_2026.tex
\newcommand{\rh}[1]{\rotatebox{90}{#1}} 
\begin{table*}[t]

\centering
\footnotesize
\setlength{\tabcolsep}{2pt}
\renewcommand{\arraystretch}{1.05}
\resizebox{\textwidth}{!}{
\begin{tabular}{l|ccccccccccc|cccc|c}
\multirow{2}{*}{Method} 
& \multicolumn{11}{c|}{Zero-Shot Classification}
& \multicolumn{4}{c|}{Image-Text Retrieval}
& \multicolumn{1}{c}{\makecell{Visual\\Centric}} \\
\cline{2-17}
& \rh{Food101} & \rh{CIFAR10} & \rh{CIFAR100} & \rh{SUN397} & \rh{Cars}
& \rh{Aircraft} & \rh{DTD} & \rh{Pets} & \rh{Cal101} & \rh{Flowers} & \rh{IN-1K}
& \rh{COCO I2T~} & \rh{COCO T2I~} & \rh{Flickr I2T} & \rh{Flickr T2I}
& \rh{MMVP} \\
\hline
\multicolumn{17}{c}{ViT-Base} \\
\hline
CLIP 
& 85.5 & 93.0 & 71.7 & 66.8 & 83.5 & 16.7 & 52.8 & 90.1 & 91.2 & 63.9 & 67.0
& 55.4 & 38.3 & 83.2 & 65.5 & 19.3 \\

DreamLIP
& 58.3 & 87.3 & 62.6 & 54.3 & 29.7 & 4.9 & 29.2 & 60.3 & 83.1 & 28.9 & 50.3
& 53.3 & 41.2 & 82.3 & 66.6 & 24.0 \\

LiT
& - & - & - & - & 13.2 & 5.0 & - & 74.4 & - & 48.2 & 56.2
& 30.0 & 16.5 & 54.8 & 38.5 & - \\

ShareLock
& - & - & - & - & 11.5 & 8.3 & - & 66.6 & - & 48.8 & 59.1
& 26.0 & 13.5 & 53.9 & 34.9 & - \\

SAIL-NV2
& 77.7 & 93.8 & 79.9 & 66.2 & 35.8 & 13.4 & \textbf{61.5} & 81.7 & 82.1 & 61.5 & 68.1
& 57.3 & \textbf{45.3} & 84.1 & \textbf{70.1} & 24.4 \\

CLIP-FT
& 81.4 & 89.6 & 72.6 & 63.3 & 81.5 & 12.6 & 49.9 & 85.7 & 88.6 & 61.8 & 64.5
& 53.8 & 37.6 & 82.1 & 62.3 & 17.0 \\

w/o LG
& 82.3 & 90.1 & 71.8 & 62.5 & 80.4 & 10.5 & 47.6 & 86.1 & 87.5 & 62.2 & 64.2
& 54.9 & 39.2 & 82.3 & 66.1 & 19.3 \\

{\methodname}
& \textbf{87.6} & \textbf{94.5} & \textbf{81.3} & \textbf{67.9} & \textbf{85.8} & \textbf{25.5} & 55.4 & \textbf{91.8} & \textbf{92.2} & \textbf{70.0} & \textbf{70.2}
& \textbf{58.8} & 39.8 & \textbf{84.8} & 67.8 & \textbf{26.0} \\
\hline

\multicolumn{17}{c}{ViT-Large} \\
\hline
CLIP
& 90.1 & 94.6 & 77.4 & 72.6 & 89.6 & 25.0 & 60.4 & 91.7 & 82.1 & 75.5 & 72.7
& 59.7 & 43.0 & 87.6 & 70.2 & 20.0 \\

SAIL-NV2
& 81.9 & 96.1 & 85.2 & 68.3 & 42.9 & 16.3 & 60.4 & 84.7 & 82.4 & 67.5 & 72.1
& 57.3 & \textbf{45.3} & 84.9 & \textbf{73.0} & 28.0 \\

CLIP-FT
& 86.5 & 92.3 & 75.7 & 68.8 & 88.3 & 19.9 & 59.0 & 90.2 & 85.4 & 72.2 & 69.3
& 55.6 & 42.7 & 82.5 & 68.9 & 19.3 \\

w/o LG
& 87.1 & 90.9 & 77.4 & 71.6 & 90.1 & 19.2 & 61.2 & 89.0 & 81.5 & 73.4 & 70.2
& 56.6 & 43.5 & 82.1 & 70.3 & 20.0 \\

{\methodname}
& \textbf{92.5} & \textbf{97.2} & \textbf{87.3} & \textbf{73.8} & \textbf{92.0} & \textbf{37.5} & \textbf{63.5} & \textbf{92.7} & \textbf{93.8} & \textbf{80.6} & \textbf{76.5}
& \textbf{64.0} & 44.4 & \textbf{89.9} & 72.4 & \textbf{29.6} \\
\hline
\end{tabular}
}
\caption{Performance on zero-shot classification, image-text retrieval, and visual-centric tasks. All scores are reported in percent. LG denotes the Language Guider.}
\label{tab:shadowclip_results}
\end{table*}

\textbf{Setup.} We train \methodname{} on CC12M with image-text contrastive loss plus relational distillation, freezing the CLIP, DINO, and LLM backbones and caching their representations once over the training set. We evaluate on zero-shot classification (11 benchmarks), image-text retrieval (MSCOCO, Flickr30k), and visual-centric understanding (MMVP), comparing against CC12M-trained baselines: CLIP, DreamLIP, LiT, ShareLock, and SAIL, all of which are trained without additional downstream-task training data or supervision. App.~\ref{app:shadowclip} gives full training and evaluation details.

\textbf{Compute efficiency.} Caching is particularly beneficial when the same representations are reused across epochs, training runs, target models, or hyperparameter sweeps. \methodname{} requires $95\%$ fewer trainable parameters  than full fine-tuning of CLIP-Large (20M vs.\ 400M), and its cached representations eliminate repeated forward and backward propagation through the CLIP backbone. This reduces its training-phase FLOPs by a factor of $2{\times}10^{3}$; including the one-time precomputation, total FLOPs over 3 training epochs is $9\times$ less.

\textbf{Results.} Table~\ref{tab:shadowclip_results} reports \methodname's performance across the three evaluation suites. \emph{Zero-shot classification.} \methodname{} generally outperforms CC12M-trained baselines across 11 benchmarks at both ViT-Base and ViT-Large, with a gain of 3.6 percentage points (pp) in average accuracy and a 3.2~pp gain on ImageNet-1K over the CLIP-Base backbone across all datasets. For \emph{visual-centric understanding:} \methodname{} improves on CLIP-Base by 6.7~pp on MMVP (19.3 to 26.0) and on CLIP-Large by 9.6~pp (20.0 to 29.6), consistent with the SwiGLU vision residual importing DINO's fine-grained visual structure. For \emph{retrieval:} \methodname{} leads on image-to-text retrieval (Flickr30k R@1: 89.9 at Large, 84.8 at Base) and is competitive on text-to-image. We note that T2I performance against SAIL is lower by construction, since SAIL replaces the CLIP text encoder with a full LLM at inference, whereas \methodname{} retains it for inference efficiency. 

% \textbf{Ablations.} Fine-tuning CLIP's linear projections directly on CC12M (CLIP-FT) \emph{degrades} performance on all three suites, since CC12M is too small to overcome dataset-specific bias. Replacing the Language Guider with fine-tuning only the linear text projection (w/o LG) eliminates most of the language-side gains. Both ablations match the geometric prediction: capability transfer must operate by adding relational structure on top of a frozen backbone, not by modifying a backbone's absolute representations.
% Overall, \methodname{} shows that capability transfer via a cached forward pass can exceed projection fine-tuning, at less compute than full fine-tuning.

\textbf{Ablations.} We ablate two principal design choices, reporting percentage-point differences averaged over the Base and Large variants, using the mean over the 11 classification or 4 retrieval settings.

\emph{(i) Fine-tuning CLIP projections (CLIP-FT).} We replace \methodname's residual modules with a baseline that simply fine-tunes the CLIP visual and textual projection layers on CC12M. Fine-tuning the projections may introduce dataset-specific bias due to the relatively small scale of CC12M, reducing the generalization ability of the original CLIP model and leading to inferior performance. Specifically, compared to the original CLIP, this baseline suffers a performance drop of 2.5~pp in zero-shot classification, 2.2~pp in image-text retrieval, and 1.5~pp in visual-centric understanding. In contrast, our method yields improvements of 6.8~pp, 4.6~pp, and 9.7~pp across these respective tasks compared to this baseline, effectively reversing the decline. This suggests that direct modification of CLIP's representation space on a small dataset can degrade performance rather than enhance it.

\emph{(ii) Removing the Language Guider (w/o LG).} We retain the vision-side SwiGLU residual but remove the language-side MLP residual and relational distillation, fine-tuning only the linear projection layer. Compared to our complete \methodname{}, removing the Language Guider leads to performance declines of 6.9~pp in zero-shot classification, 3.4~pp in image-text retrieval, and 8.2~pp in visual-centric understanding. The pronounced drop in visual-centric tasks suggests that the model fails to effectively internalize the fine-grained perceptual capabilities inherited from DINO. Without the Language Guider, simply fine-tuning the linear projection layer of CLIP is insufficient for the language side to interpret the additional visual information introduced by DINO. The original balance between the vision and language is disrupted, which supports the view that capability transfer on the language side requires the relational-distillation operator predicted by §\ref{sec:training_dynamics}, not merely linear adaptation.

\textbf{Summary.} \methodname{} demonstrates the potential of Shadow Casting: capability transfer via a cached forward pass can outperform projection fine-tuning at less compute than full fine-tuning.  %%
% !TEX root = ../neurips_2026.tex

\section{Limitations and Open Questions}
\label{sec:openquestions}
Our results establish that directional structure is shared across models, but several questions remain. 
(i) \textbf{Scale}: our analysis spans 44 models across vision and language, but does not yet reach the largest frontier systems. Whether displacement parallelism strengthens or degrades at extreme scale remains unknown.
% (ii) \textbf{Proximity measures}: we do not compute mutual-kNN or CKA on these evaluations and thus cannot directly test whether directional agreement persists where such measures weaken.
(ii) \textbf{Proximity measures}: we do not evaluate mutual-kNN or CKA here, leaving open whether directional agreement persists where these measures weaken.
(iii) \textbf{Uniqueness}: we have shown that independently trained models converge on a shared directional geometry, but not that this geometry is the \emph{only} one compatible with high performance. The shared structure we observe may be one of several equally viable modes, or it may be unique among strong models, what \citet{huh2024position} term the ``Anna Karenina'' scenario. Our evidence is consistent with the former but does not exclude the latter. 
(iv) \textbf{Locality}: we deliberately use the simplest directional comparison---cosine similarity of difference vectors under rigid alignment---but other approaches are possible. A single global orthogonal map cannot recover shared structure requiring region-specific rotations, anisotropic scaling, or nonlinear transformations, so low similarity under this constraint need not imply its absence. Relation-conditioned sampling, such as within-class, sibling-concept, or semantically distant pairs, and richer decompositions across layers and functional subspaces could reveal finer structure obscured by our global comparison. 
(v) \textbf{Human alignment}: strong agreement among models need not imply equally strong alignment with human conceptual organization~\citep{muttenthaler2025aligning}. Our models span diverse training objectives, so the observed convergence cannot be attributed specifically to human labels. Whether this shared geometry matches human representations remains open, motivating comparisons with human similarity judgments or neural responses.

% !TEX root = ../neurips_2026.tex
\section{Conclusion}
\label{sec:conclusion}
We have shown that independently trained models share a geometric structure that is not captured in the prevailing view. Our findings suggest that models do converge to a shared reality, and that this structure is visible in directions between samples even when their aligned positions differ. This directional structure is substantially preserved under trivial alignment across 44 vision and language encoders. It decomposes into a shared semantic component that aligns linearly and a private capability component that does not---components directly actionable for capability transfer. These results open new possibilities for model generalization and capability transfer.

\clearpage
\begin{ack}
Breuer Lab gratefully acknowledges the support of the OpenAI Cybersecurity Grant.
\end{ack}

\bibliographystyle{abbrvnat}
\bibliography{references}

\appendix
\clearpage
% !TEX root = ../neurips_2026.tex
\section{Additional Experimental Results}
\label{app:addlexperiments}
This appendix reports additional analyses that complement the main empirical results. \S\ref{app:heatmapsection} examines the displacement geometry pairwise, and  \S\ref{app:failure} reports controlled simulations that test the inheritance account of \S\ref{sec:orthogonality} by selectively breaking each of its two concept-level conditions.
\subsection{Displacement geometry is observable at the level of individual model pairs}
\label{app:heatmapsection}
Fig.~\ref{fig:heatmap-allpairs} shows the full $44 \times 44$ similarity matrix across our model set, with \textbf{Pointwise} (left) and \textbf{Cross-Model} (right) plotted on a shared colorbar at the sample level (top row) and the concept level (bottom row). Three patterns are visible directly in the matrices. First, Cross-Model (\emph{right side}) exceeds Pointwise (\emph{left side}) on every block: at both sample and concept resolution, the right-hand matrices are uniformly redder than the left, including for cross-modal (V-L) pairs. Second, within-modality blocks are the strongest sources of agreement, with the language block (lower-right) the most saturated, followed by the vision block (upper-left), and the V-L off-diagonal weakest: L-L $>$ V-V $>$ V-L, consistent with the bar-chart aggregates of \S\ref{sec:concept_main}. Third, concept-level Cross-Model dominates sample-level Cross-Model: the same model pairs that are pale at sample-level Cross-Model (Fig.~\ref{fig:heatmap-sample-cm}) are visibly redder at concept-level Cross-Model (Fig.~\ref{fig:heatmap-concept-cm}). This increase is consistent with concept centroids averaging away instance-specific variation, such as background, pose, and wording, and isolating shared semantic factors that individual samples mix with model-specific weights.

The displacement geometry is thus visible at the level of individual model pairs, not only in aggregate:

\begin{figure}[!htbp]
  \centering
  \begin{subfigure}{0.49\textwidth}
    \centering
    \includegraphics[width=\linewidth]{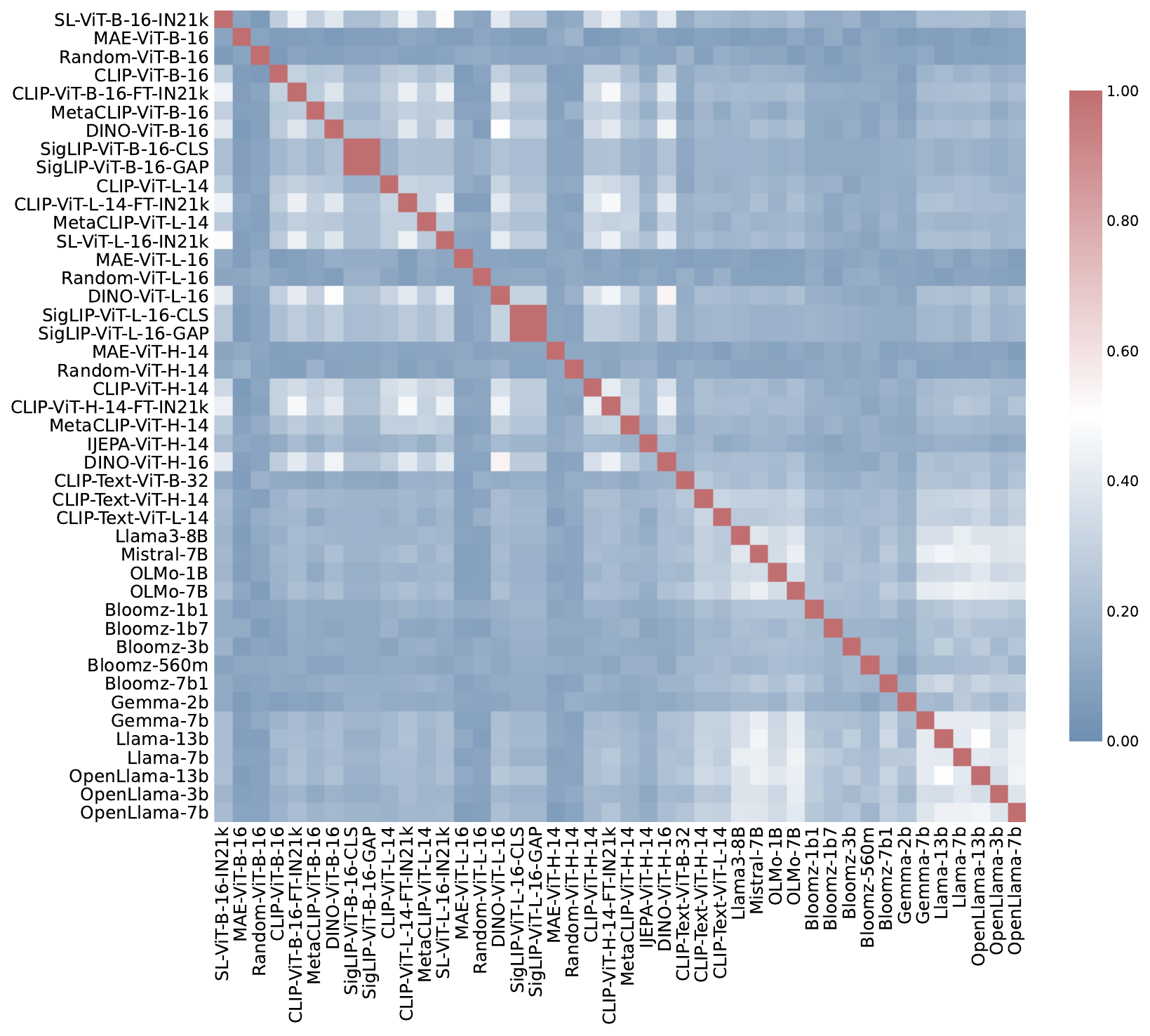}
    \caption{Pointwise, sample-level.}
    \label{fig:heatmap-sample-pw}
  \end{subfigure}
  \hfill
  \begin{subfigure}{0.49\textwidth}
    \centering
    \includegraphics[width=\linewidth]{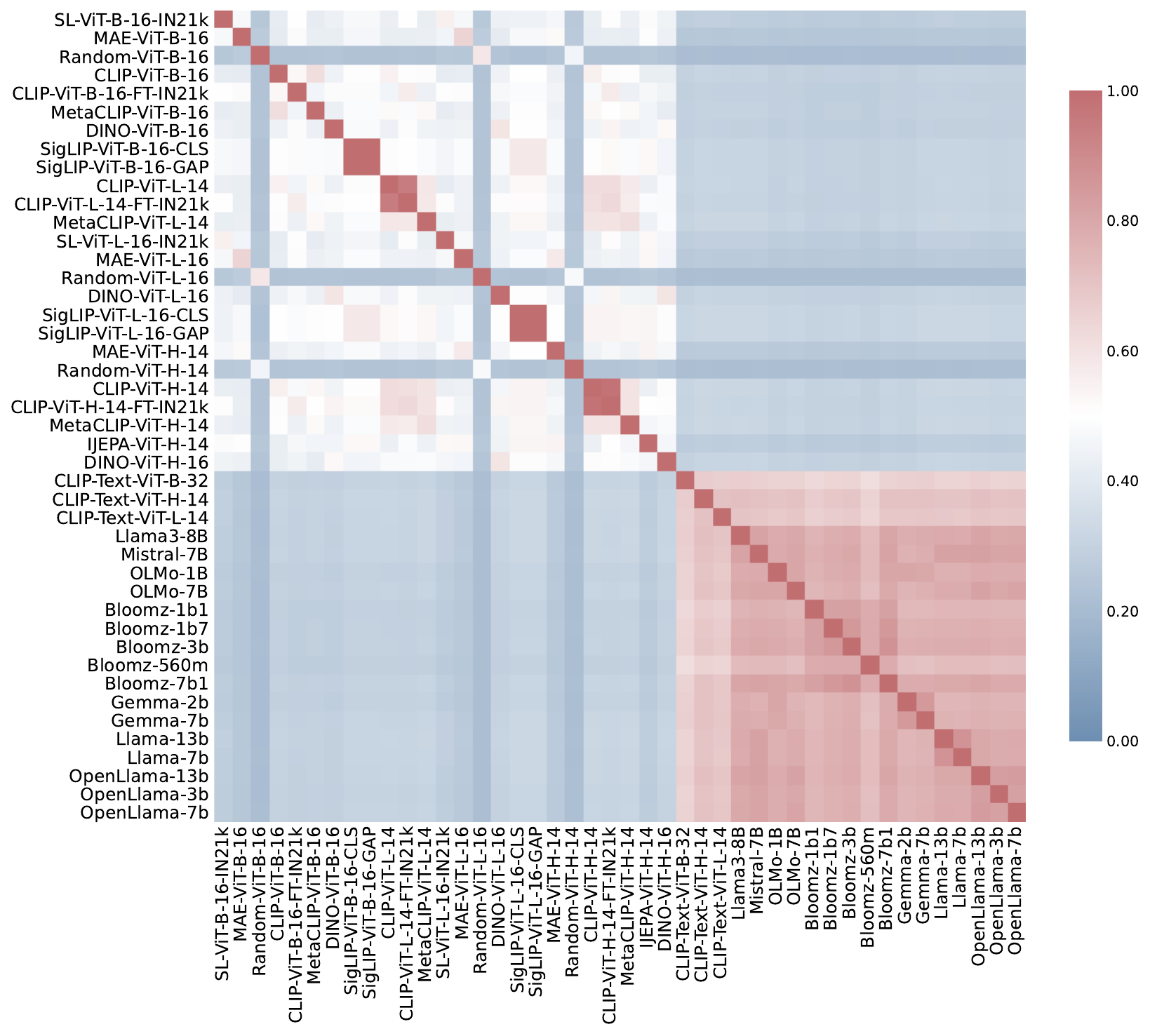}
    \caption{Cross-Model, sample-level.}
    \label{fig:heatmap-sample-cm}
  \end{subfigure}

  \begin{subfigure}{0.49\textwidth}
    \centering
    \includegraphics[width=\linewidth]{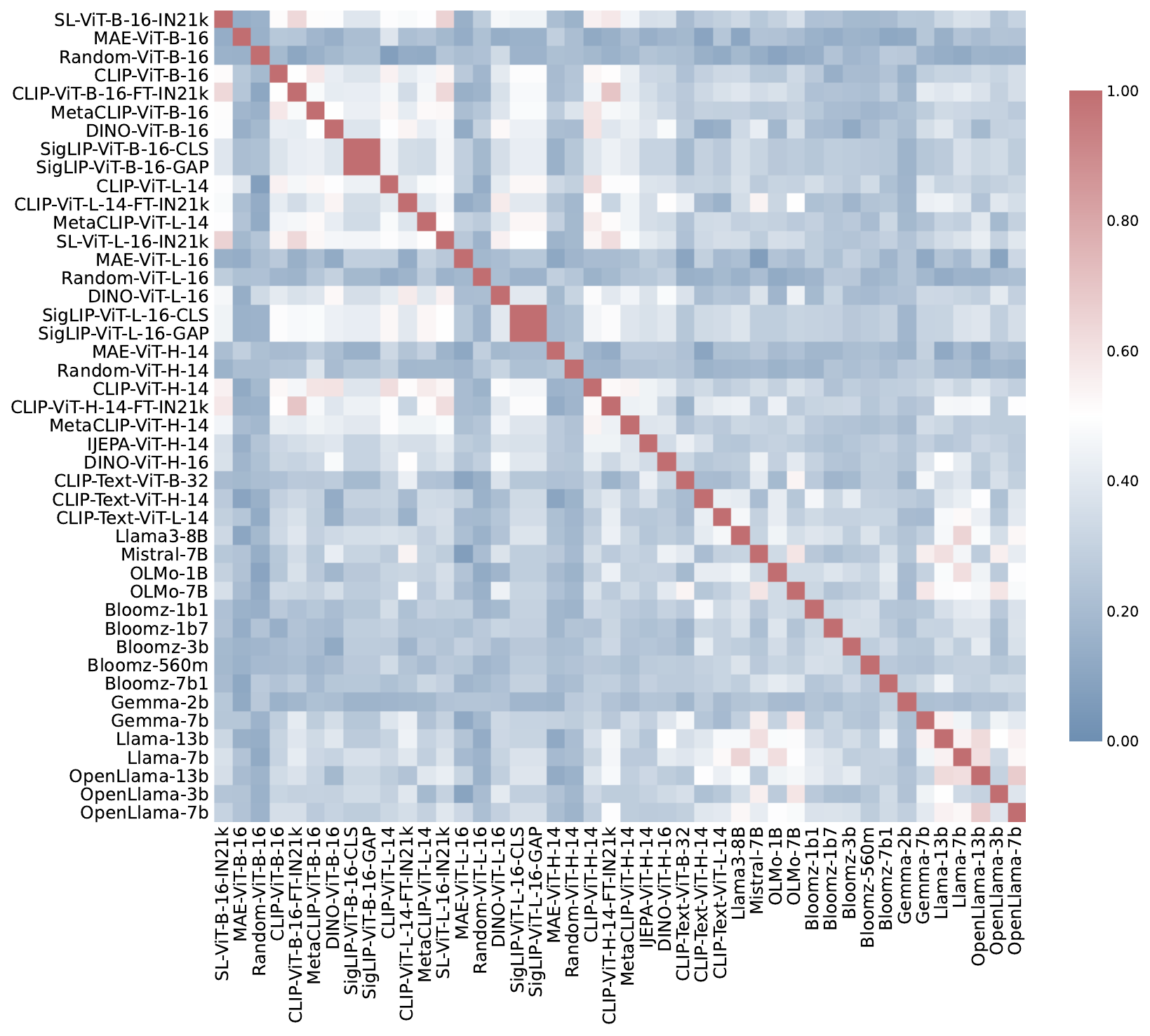}
    \caption{Pointwise, concept-level.}
    \label{fig:heatmap-concept-pw}
  \end{subfigure}
  \hfill
  \begin{subfigure}{0.49\textwidth}
    \centering
    \includegraphics[width=\linewidth]{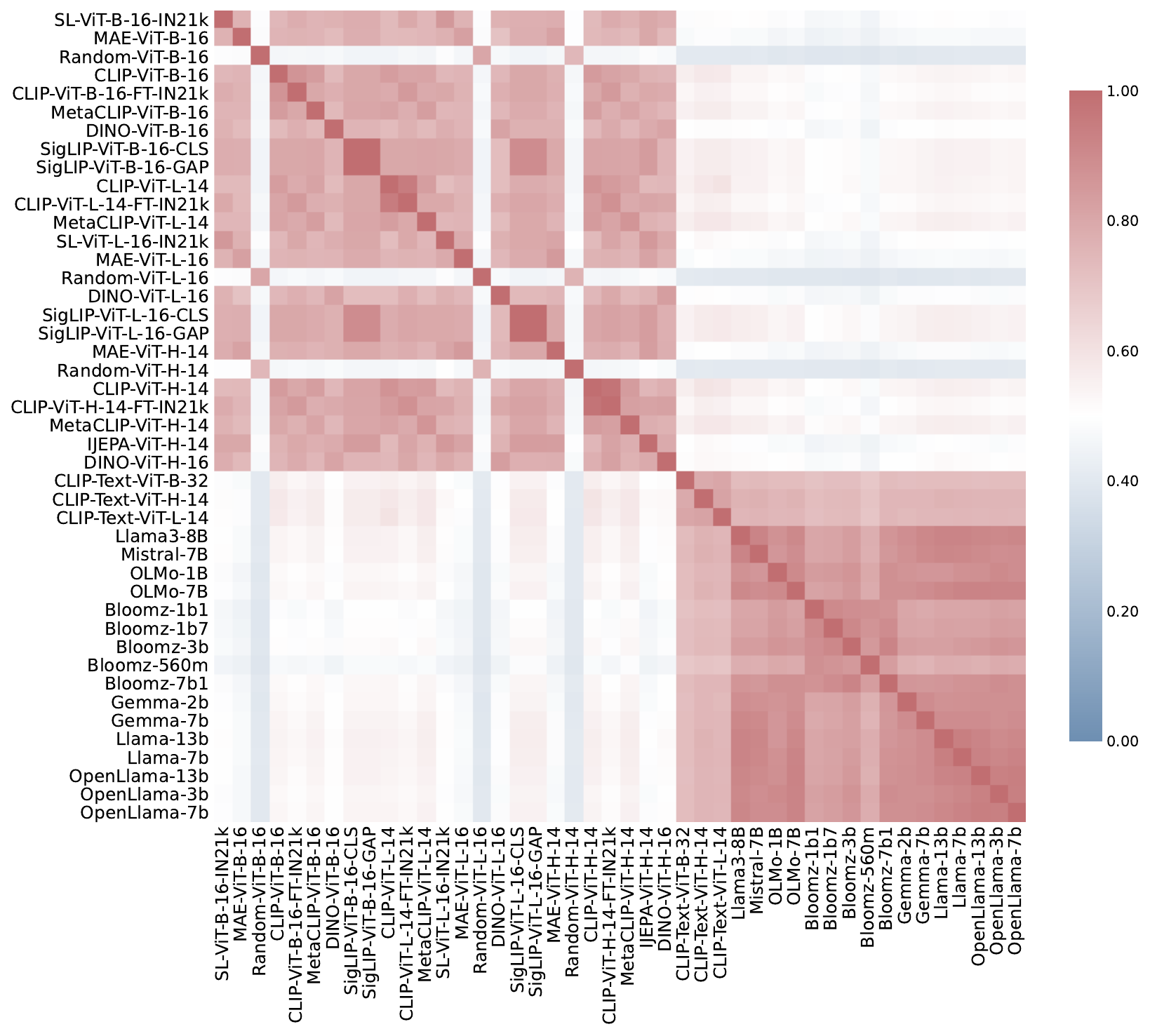}
    \caption{Cross-Model, concept-level.}
    \label{fig:heatmap-concept-cm}
  \end{subfigure}
  \caption{All-to-all similarity across the 44 models. \emph{Top row}: sample level. \emph{Bottom row}: concept level. Pointwise (\emph{left}) and Cross-Model (\emph{right}) on a shared colorbar within each row.}
  \label{fig:heatmap-allpairs}
\end{figure}

\begin{figure}[t]
  \centering
  \begin{subfigure}{0.49\textwidth}
    \centering
    \includegraphics[width=\linewidth]{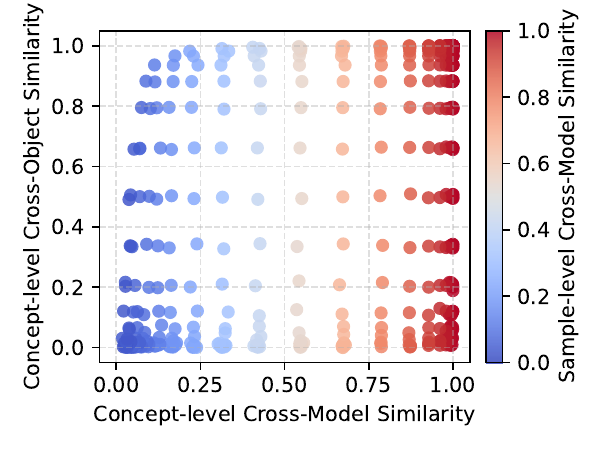}
    \caption{Changes in sample-level Cross-Model similarity\\\phantom{11.a }when concept-level \emph{parallelism} fails.}
  \end{subfigure}
  \hfill
  \begin{subfigure}{0.49\textwidth}
    \centering
    \includegraphics[width=\linewidth]{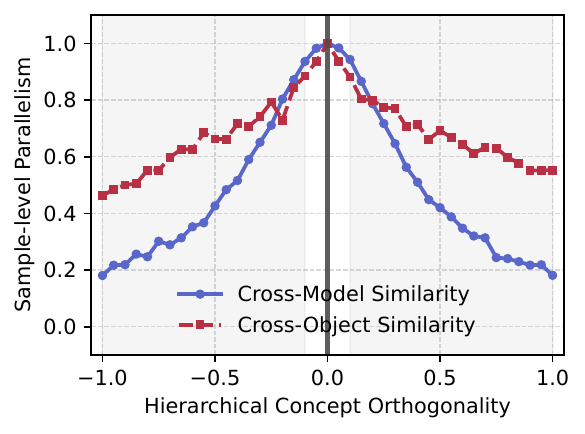}
    \caption{Changes in sample-level parallelism\\\phantom{11.a }when concept-level \emph{orthogonality} fails.}
  \end{subfigure}
  \caption{Degradation of sample-level geometry under concept-level geometric collapse.}
  \label{fig:failure}
\end{figure}

\subsection{Simulations: degrading concept-level geometry lowers sample-level parallelism}
\label{app:failure}
Section \ref{sec:orthogonality} predicts that sample-level parallelism inherits from two concept-level conditions: (i) cross-model concept parallelism  and (ii) within-model hierarchical concept orthogonality. We test the inheritance account in controlled simulation by constructing two ideally aligned representation spaces and selectively breaking each condition.

\textbf{Setup.} We embed a three-level hierarchy of $4+16+64=84$ concepts into two $d=512$ spaces, initially satisfying both (i) and (ii). Sample representations are random weighted combinations of concepts: arbitrary subsets in Sim.~1, root-to-leaf paths in Sim.~2.

\textbf{Sim.~1: breaking concept-level parallelism.} We sweep concept-level Cross-Model and Cross-Object similarity independently and measure sample-level Cross-Model. As predicted by (i), sample-level Cross-Model tracks concept-level Cross-Model, not Cross-Object (Fig.~\ref{fig:failure}, left): the color gradient is aligned with the x-axis, and concept-level Cross-Object varies independently of sample-level outcomes. Failure of concept-level parallelism therefore propagates: when concepts are not parallel across models, the displacement directions between samples are not either.

\textbf{Sim.~2: breaking within-model orthogonality.} We vary the cosine between child-concept increments and their parent vector, moving the system away from the orthogonal regime where (ii) holds. We then form samples along root-to-leaf paths and measure sample-level Cross-Model and Cross-Object similarity. Both decay as orthogonality fails (Fig.~\ref{fig:failure}, right), with Cross-Model decaying faster than Cross-Object. In this construction, sample-level parallelism peaks at zero parent-child cosine.

Within this generative model, sample-level parallelism follows concept-level geometry and drops when either concept-level condition is broken.

\subsection{Asymmetric pairs}
\label{app:asym}

\begin{table}[htbp]
\centering
\small
\caption{Similarity between asymmetric model pairs.}
\label{tab:model_pair_similarity}
\begin{tabular}{cc|ccc}
\toprule
\textbf{Model 1} & \textbf{Model 2} & \textbf{Pointwise} & \textbf{Cross-Object} & \textbf{Cross-Model} \\
\midrule
CLIP-ViT-Huge & MAE-ViT-Base  & 0.059 & 0.996 & 0.376 \\
Llama3-8B     & Bloomz-560M   & 0.182 & 0.994 & 0.611 \\
Llama3-8B     & CLIP-ViT-Huge & 0.204 & 0.965 & 0.306 \\
\bottomrule
\end{tabular}
\end{table}

\subsection{Concept geometry across modalities, scales, and reference choices}
\label{app:concept-orthogonality}
\begin{figure}[t]
  \centering
  \includegraphics[width=0.4\textwidth]{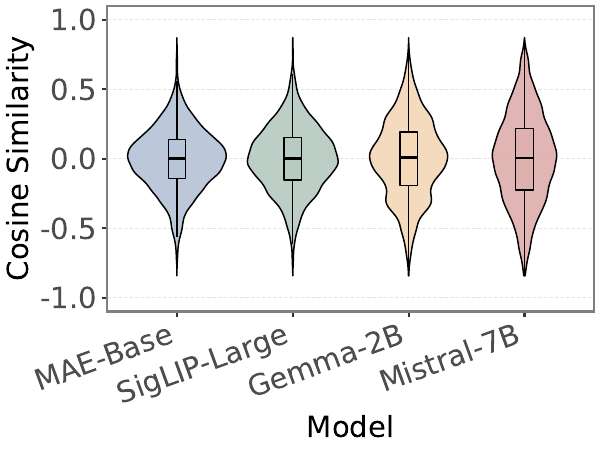}
  \caption{Hierarchical concepts orthogonality of different models across modalities.}
  \label{fig:robustness}
\end{figure}

\textbf{Orthogonality across modalities and scales.} Fig.~\ref{fig:robustness} plots the per-pair distribution of $\cos(\boldsymbol{v}_{\text{child}} - \boldsymbol{v}_{\text{parent}},\; \boldsymbol{v}_{\text{parent}})$ averaged over all layers, for four models spanning vision (MAE-Base, SigLIP-Large) and language (Gemma-2B, Mistral-7B). All four distributions are symmetric and centered at zero: parent concepts and child displacements are orthogonal on average. Orthogonality is therefore prevalent across models of different modalities and scales, with vision models concentrated slightly tighter around zero than language models.

\textbf{Random-reference control.} Table~\ref{tab:concept_reference_control} compares a parent concept with a random unrelated concept as the reference. Let $\boldsymbol{v}_{\text{child}}$ and $\boldsymbol{v}_{\text{sibling}}$ denote the representations of sibling child concepts, and $\boldsymbol{v}_{\text{ref}}$ the reference representation. The parent reference yields positive child--reference similarity and orthogonality between the child--reference displacement and the reference, whereas the random reference does not, indicating that these relationships are not generic properties of arbitrary concept pairs.

\begin{table}[htbp]
\centering
\small
\caption{Concept geometry with random unrelated and parent references.}
\label{tab:concept_reference_control}
\begin{tabular}{c|ccc}
\toprule
\textbf{Reference} & $\cos(\boldsymbol{v}_{\text{child}},\boldsymbol{v}_{\text{ref}})$ & $\cos(\boldsymbol{v}_{\text{child}}-\boldsymbol{v}_{\text{ref}},\boldsymbol{v}_{\text{ref}})$ & $\cos(\boldsymbol{v}_{\text{child}}-\boldsymbol{v}_{\text{ref}},\boldsymbol{v}_{\text{sibling}}-\boldsymbol{v}_{\text{ref}})$ \\
\midrule
Random & 0.00 & -0.71 & 0.73 \\
Parent & 0.72 & 0.00 & -0.30 \\
\bottomrule
\end{tabular}
\end{table}

\subsection{Effect of alignment flexibility on measured similarity}
\label{app:alignment_flexibility}

We compare orthogonal Procrustes alignment (OPA), OPA with global scaling, OPA with per-dimension scaling, and full linear alignment on paired whitened 64-dimensional representations with preset shared correlations $\rho \in \{0.1, 0.2, \ldots, 0.9\}$. Each alignment is fitted on 50,000 training points and evaluated on 20,000 displacement pairs drawn from a disjoint pool of 10,000 test points, with results averaged over 10 seeds. Table~\ref{tab:alignment_flexibility_results} shows that OPA closely recovers the preset correlation, with a mean signed error of $-0.004$, and global scaling leaves its scores unchanged. Per-dimension scaling and full linear alignment instead exceed the preset correlation by $0.038$ and $0.063$ on average, respectively, showing that more flexible alignment can increase measured similarity beyond the preset shared correlation in this synthetic setting.

\begin{table}[htbp]
\centering
\small
\caption{Held-out displacement similarity under different alignment methods. Mean error is the signed difference between the measured score and the preset correlation $\rho$.}
\label{tab:alignment_flexibility_results}
\begin{tabular}{c|cccc}
\toprule
\textbf{Preset $\rho$} & \textbf{OPA} & \textbf{OPA + global scale} & \textbf{OPA + per-dimension scaling} & \textbf{Full linear} \\
\midrule
0.1 & 0.093 & 0.093 & 0.131 & 0.151 \\
0.2 & 0.193 & 0.193 & 0.252 & 0.285 \\
0.3 & 0.295 & 0.295 & 0.361 & 0.399 \\
0.4 & 0.396 & 0.396 & 0.460 & 0.498 \\
0.5 & 0.496 & 0.496 & 0.553 & 0.588 \\
0.6 & 0.596 & 0.596 & 0.641 & 0.670 \\
0.7 & 0.697 & 0.697 & 0.728 & 0.748 \\
0.8 & 0.798 & 0.798 & 0.814 & 0.826 \\
0.9 & 0.899 & 0.899 & 0.904 & 0.907 \\
\midrule
Mean error (score $-\rho$) & $-0.004$ & $-0.004$ & $+0.038$ & $+0.063$ \\
\bottomrule
\end{tabular}
\end{table}

% !TEX root = ../neurips_2026.tex

\section{Statistical Significance Tests}
\label{app:statistics}
Paired Wilcoxon signed-rank tests confirm the statistical significance of the findings in \S\ref{sec:concept_main} and \S\ref{ssec:samplelevel}. As shown in Table~\ref{tab:wilcoxon}, at both sample and concept levels, and in all three modality groupings, Cross-Model exceeds Pointwise by a median that is significantly positive (all six tests $p < 0.0001$).
\begin{table}[h]
\footnotesize
\caption{Wilcoxon signed-rank tests of Cross-Model $>$ Pointwise across modality groups, at sample and concept levels. Median $D$ is the median paired difference $D = \text{Cross-Model} - \text{Pointwise}$, with bootstrap standard error; $p$-values are Benjamini--Hochberg adjusted across all six tests.}
\centering
\begin{tabular}[t]{ccccc}
\toprule
Modality Group & Sample Median D & Sample p-value & Concept Median D & Concept p-value\\
\midrule
V-V & $0.231 \pm 0.008$ & $<\!0.0001$ & $0.407 \pm 0.011$ & $<\!0.0001$\\
L-L & $0.387 \pm 0.007$ & $<\!0.0001$ & $0.472 \pm 0.011$ & $<\!0.0001$\\
V-L & $0.146 \pm 0.002$ & $<\!0.0001$ & $0.242 \pm 0.003$ & $<\!0.0001$\\
\bottomrule
\end{tabular}
\label{tab:wilcoxon}
\end{table}
% !TEX root = ../neurips_2026.tex
\section{Deferred Details of Preprocessing, Alignment, and Decomposition}
\label{app:prelim}

This section provides derivations and details deferred from §\ref{sec:prelim}.

\textbf{PCA dimensionality.} Across our 44 models, $d_m$ ranges from a few hundred to several thousand. Cross-model comparisons must therefore be made under a common dimensional constraint, since representational quality is itself dimension-dependent: a model given more dimensions to work with has more room to encode redundancy or structure that another model lacks. We project each $X_m$ to a common dimension $d$ via PCA, which both matches dimensionalities and removes a portion of the per-axis noise. We report results at $d \in \{20, 100, 500\}$ and find the qualitative findings stable across this range, as shown in Table~\ref{tab:pca}.

\begin{table}[h]
\centering
\footnotesize
\caption{Comparison: Pointwise vs. Cross-Model similarity across different PCA dimensions.}
\begin{tabular}{ccccccc}
\toprule
\multirow{2}{*}{Similarity} 
& \multicolumn{2}{c}{PCA-20} 
& \multicolumn{2}{c}{PCA-100} 
& \multicolumn{2}{c}{PCA-500} \\
\cmidrule(lr){2-3} \cmidrule(lr){4-5} \cmidrule(lr){6-7}
& Pointwise & Cross-Model 
& Pointwise & Cross-Model 
& Pointwise & Cross-Model \\
\midrule
V-V 
& $0.50 \pm 0.11$ & $0.69 \pm 0.09$ 
& $0.37 \pm 0.13$ & $0.61 \pm 0.09$ 
& $0.31 \pm 0.13$ & $0.45 \pm 0.07$ \\

L-L 
& $0.53 \pm 0.09$ & $0.84 \pm 0.08$ 
& $0.52 \pm 0.10$ & $0.84 \pm 0.06$ 
& $0.36 \pm 0.08$ & $0.73 \pm 0.07$ \\

V-L 
& $0.40 \pm 0.09$ & $0.43 \pm 0.05$ 
& $0.24 \pm 0.04$ & $0.32 \pm 0.04$ 
& $0.19 \pm 0.04$ & $0.31 \pm 0.02$ \\
\bottomrule
\end{tabular}
\label{tab:pca}
\end{table}

As the PCA dimensionality decreases, all similarity metrics increase, with Cross-Model similarity consistently exceeding Pointwise similarity. To retain as much information as possible, results for other experiments are reported using 500 PCA dimensions.

\textbf{Whitening.} Following the linear representation hypothesis~\citep{park2023linear}, we argue that for geometric metrics to have clear semantic meaning, representations should ideally reside in a causal inner-product space. However, standard training objectives in modern vision and language models constrain the model's outputs but not its internal coordinate system. As a consequence, the raw representation space of a trained model is typically anisotropic: variance differs sharply along different axes, so cosine similarity in the raw space conflates two unrelated quantities, the angle between underlying semantic directions and the relative scaling of those directions in the model's coordinate system. In such spaces, the standard Euclidean inner product fails to reliably reflect underlying semantic relationships~\citep{uselis2026compositional}.

As a simple approximation to such a space, we apply a whitening transformation to each model's representations independently.
Concretely, given the centered representation $X_m - \mathbf{1}\boldsymbol{\mu}_m^\top$ with empirical covariance $\Sigma_m$, the whitening transform is $W_m = \Sigma_m^{-1/2}$ and the whitened representation is
\begin{equation}
    \widehat{X}_m = (X_m - \mathbf{1}\boldsymbol{\mu}_m^\top)\, W_m.
\end{equation}
This follows \citet{park2023linear}, who estimate the causal inner product by whitening with the representation covariance. After whitening, the Euclidean inner product on $\widehat{X}_m$ is what cosine similarity in the original space would compute if the original space were isotropic. Subsequent geometric measures, including all three metrics in §\ref{sec:prelim} and the decomposition below, are computed on the whitened representations.

\begin{figure}[t]
  \centering
  \includegraphics[width=0.8\textwidth]{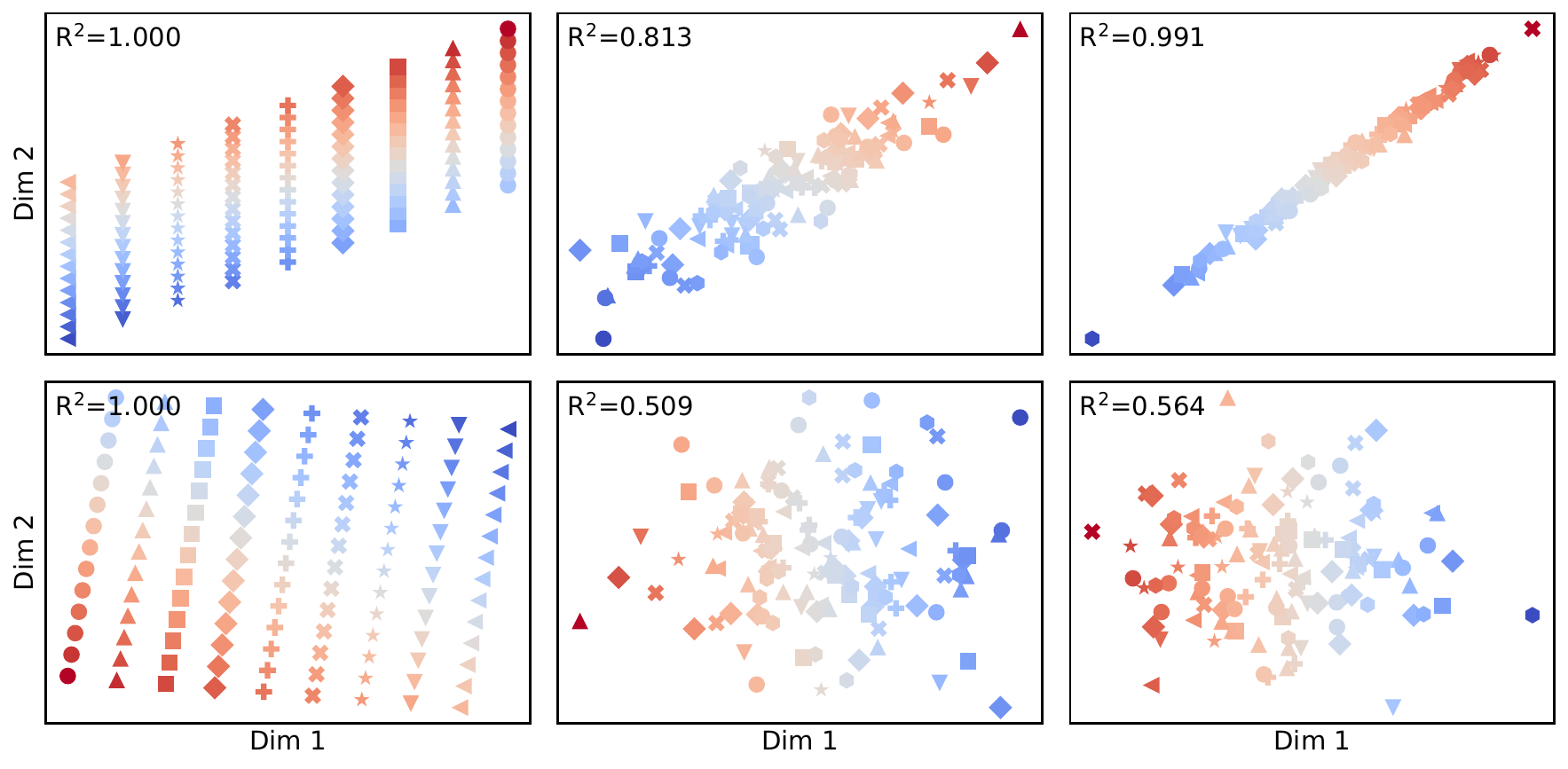}
  \caption{Influence of whitening on representation space.}
  \label{fig:whiten}
\end{figure}

\textbf{Empirical illustration of the necessity of whitening.} To intuitively demonstrate how anisotropy leads to the non-identifiability of semantic structure, we examine three representative cases in a hypothetical two-dimensional representation space, as shown in Fig.~\ref{fig:whiten}. The top row shows representations in the original space, the bottom row shows representations after whitening, while the three columns from left to right correspond to the three distinct scenarios. These cases highlight why the raw Euclidean metric can be deceptive and how whitening recovers the true geometric relationships. $R^2$ is fraction of explained variance by additive reconstruction (sum of per-concept effects) on a balanced $8\times8$ concept grid.
\begin{itemize}
    \item \textit{Left:} Structurally Sound Representation. When the representation is already linearly factored (i.e., each embedding is a pure sum of concept components without noise), the $R^2$ score is $1.0$. In this ideal scenario, the space is naturally aligned with the causal inner product, and whitening preserves this perfect score, confirming that the transformation does not distort already-well-structured data.
    \item \textit{Middle:} Noise Dominance and Spurious Linearity. In this scenario, the representation is only partially linear, with independent noise $\epsilon_{ij}$ dominating the overall variance. However, because the scale of the noise is smaller than that of the primary concept, the raw space yields a deceptively high $R^2$ of $0.813$. Whitening rescales the dominant direction so that the noisy direction counts equally, and the score drops to $0.509$.
    \item \textit{Right:} Dimensional Collapse and Missing Concepts. In the most extreme case, the representation lacks any information regarding the second concept. Nevertheless, due to the severe anisotropy of the raw space, the $R^2$ remains misleadingly high at $0.991$. After whitening gives the nearly empty second direction equal weight, the score falls to $0.564$.
\end{itemize}
These examples collectively demonstrate that in an anisotropic raw space, a high similarity or $R^2$ score may merely be a statistical artifact of dominant variance rather than a reflection of genuine semantic factorization. Whitening reduces this effect by weighting all directions equally.

\textbf{Procrustes alignment.} For each ordered pair of models $(a, b)$, we align their whitened representations on a held-out subset of inputs and evaluate metrics on the disjoint complement. The alignment solves
\begin{equation}
    R_{ab} = \arg\min_{R^\top R = I}\, \big\|\, \widehat{X}_a R - \widehat{X}_b\, \big\|_F,
\end{equation}
which has a closed-form solution via the SVD of $\widehat{X}_a^\top \widehat{X}_b$. The alignment is therefore rigid: rotation and reflection only, with no scaling or additional learned mapping. For this pairwise comparison, the aligned representations are $Z_a=\widehat{X}_aR_{ab}$ and $Z_b=\widehat{X}_b$. We  reduce the models to a consistent dimensionality, such as 20, 100, or 500. To ensure a fair setup for both sample-level and concept-level experiments, we fit the Procrustes alignment matrix using 10,000 matched items: image-caption pairs from Flickr30k at the sample level, or concept centroids from the intersection of ImageNet-22K and CommonWords79k at the concept level. We then randomly sample another 10,000 matched items from the remaining dataset for testing. PCA, whitening, and Procrustes alignment are fitted only on the alignment split and kept fixed during evaluation. The split is random over matched image-caption pairs at the sample level and concept centroids at the concept level, rather than over parent-concept groups. This ensures that the number of constraints far exceeds the degrees of freedom, satisfying the requirements for statistical reliability. Pointwise is computed over individual held-out items, whereas Cross-Object and Cross-Model are computed over 10,000 unordered pairs sampled uniformly from the same evaluation pool. Appendix~\ref{app:alignment_flexibility} compares rigid and more flexible alignment methods in a controlled synthetic setting.

\textbf{Mean restoration.} After alignment, we translate each model representations by $\boldsymbol{\mu}_m P_m W_m$ i.e. its mean mapped through PCA projection $P_m$ and whitening $W_m$ to retain the position of the original representation space. This is identical for all inputs of a model. It cancels in Cross-Model displacements but factors into Pointwise and Cross-Object. When two models' means differ, Pointwise drops and Cross-Object rises.

\textbf{Cross-model consensus decomposition.} For the decomposition and inverse transform below, $X_m \in \mathbb{R}^{n\times d}$ denotes the input feature matrix before whitening. For decomposition, we instead jointly align all models using GPA. All $R_m$ are initialized to $I$ and iteratively updated; no model is held fixed as a reference. The aligned representations are $Z_m=\widehat{X}_mR_m$. To recover the shared semantic and private capability subspaces claimed in §\ref{sec:intro}, we form a cross-model agreement matrix that scores directions by how consistently models respond to the same inputs. Averaging symmetrized cross-covariances over all $M(M-1)$ ordered model pairs gives
\begin{equation}
    S_{\mathrm{cross}} = \frac{1}{M(M-1)} \sum_{a \neq b} \frac{Z_a^\top Z_b + Z_b^\top Z_a}{2(n-1)}.
    \label{eq:scross}
\end{equation}
$S_{\mathrm{cross}}$ is symmetric by construction. We eigendecompose it and sort the eigenvectors by decreasing eigenvalue. For a chosen rank $r$, the top-$r$ eigenvectors define the consensus (shared) subspace and the bottom-$r$ eigenvectors define the low-consensus (private) subspace. Let $P_r^{\mathrm{com}}, P_r^{\mathrm{priv}} \in \mathbb{R}^{d \times r}$ be the corresponding orthonormal basis matrices; the decomposed components are
\begin{equation}
    Z_m^{\mathrm{com}}(r) = Z_m P_r^{\mathrm{com}},
    \qquad
    Z_m^{\mathrm{priv}}(r) = Z_m P_r^{\mathrm{priv}}.
\end{equation}
The decomposition is label-free: $S_{\mathrm{cross}}$ is computed from representations only, and labels are used only for downstream evaluation.

\textbf{Inverse transform for downstream evaluation.} Some downstream evaluations (e.g., linear probing or zero-shot classification heads in §\ref{sec:decomp}) require the decomposed components in each model's pre-whitening feature space rather than in the aligned, whitened space. The decomposition is performed in the aligned-whitened space $Z_m$, so we first reconstruct each component in this space via $(P_r^q)^\top$, then rotate back via $R_m^\top$, undo whitening via $W_m^{-1}$, and uncenter. For component $q \in \{\mathrm{com}, \mathrm{priv}\}$,
\begin{equation}
    X_m^{q}(r) = Z_m^{q}(r)\,(P_r^q)^\top R_m^\top W_m^{-1} + \mathbf{1}\boldsymbol{\mu}_m^\top.
\end{equation}
$X_m^{q}(r)$ lies in the same space as the pre-whitening representation $X_m$, so any classifier or zero-shot head defined on $X_m$ applies directly.

\phantomsection
\label{app:hsic}
\textbf{Statistical dependence between shared and private components.}
We measure dependence between the shared and private components using normalized Hilbert--Schmidt Independence Criterion (HSIC)~\citep{gretton2005measuring}. For each model, we compute the shared and private coordinates in the aligned, whitened space:
\[
    U = ZP_r^{\mathrm{com}}, \qquad V = ZP_r^{\mathrm{priv}},
\]
without applying the inverse transform described above. Let $\boldsymbol{u}_i$ and $\boldsymbol{v}_i$ denote corresponding rows of $U$ and $V$. We construct Gram matrices $K_{ij}=k(\boldsymbol{u}_i,\boldsymbol{u}_j)$ and $L_{ij}=l(\boldsymbol{v}_i,\boldsymbol{v}_j)$, and center them as
\[
    H = I_n - \frac{1}{n}\mathbf{1}\mathbf{1}^{\top},
    \qquad K_{\mathrm{c}} = HKH, \qquad L_{\mathrm{c}} = HLH.
\]
The normalized dependence score is
\[
    \operatorname{nHSIC}(U,V) =
    \frac{\langle K_{\mathrm{c}},L_{\mathrm{c}}\rangle_F}
    {\sqrt{\max\!\left(\|K_{\mathrm{c}}\|_F^2\|L_{\mathrm{c}}\|_F^2,\epsilon\right)}},
    \qquad \epsilon = 10^{-12}.
\]
We use linear and RBF kernels:
\[
    k_{\mathrm{linear}}(\boldsymbol{x},\boldsymbol{x}') = \boldsymbol{x}^\top\boldsymbol{x}',
    \qquad
    k_{\mathrm{RBF}}(\boldsymbol{x},\boldsymbol{x}') =
    \exp\!\left(-\frac{\|\boldsymbol{x}-\boldsymbol{x}'\|_2^2}{2\sigma^2}\right).
\]
For each component separately, $\sigma^2$ is the median of off-diagonal squared pairwise distances exceeding $\epsilon$; if no such distances exist, we set $\sigma^2=1$. The linear kernel measures linear dependence, whereas the RBF kernel can also capture nonlinear dependence.

% !TEX root = ../neurips_2026.tex
\section{Additional Details of Experiments}
\label{app:experiments}

\subsection{Training independence}
\label{app:independence}
Our paper makes claims about \emph{independently trained} models, but at the scale of contemporary foundation models, complete independence of training data is unattainable. Specifically, most modern large language models in our evaluation set (Llama, Gemma, Mistral, OLMo) draw from overlapping web-scale corpora including Common Crawl, GitHub, books, and Wikipedia. Vision encoders trained on web-scraped image-text pairs (CLIP, SigLIP, MetaCLIP) likewise share substantial overlap. We therefore use ``independently trained'' in the conventional sense established by prior work on representational alignment~\citep{huh2024position,sucholutsky2023getting}: models with distinct architectures, training objectives, optimization procedures, and data curation pipelines, but not necessarily disjoint data sources.
Three aspects of our evaluation provide evidence that the displacement parallelism we report is not solely an artifact of shared training data. First, our cross-modal (V-L) results compare vision and language models trained on fundamentally different data modalities, where direct training-data overlap is minimal even when web sources are shared. Second, our asymmetric capability pairs (App.~\ref{app:asym}) include comparisons between models trained on substantially different data scales and curation pipelines (e.g., MAE on ImageNet vs. CLIP on web-scraped image-text pairs), where shared structure is preserved despite divergent training corpora. Third, our null baseline of randomly initialized models with the same architecture but no training shows near-zero Cross-Model similarity, ruling out a purely architectural explanation for the displacement convergence we observe.

Fully disentangling shared training data from shared world structure (which is itself the underlying claim of PRH) is not straightforward at scale, and remains an open question.

% !TEX root = ../neurips_2026.tex
\section{Additional Details of \methodname{}}
\label{app:shadowclip}

\suppressfloats[t]
\begin{figure}[t]
  \centering
  \includegraphics[width=0.725\textwidth]{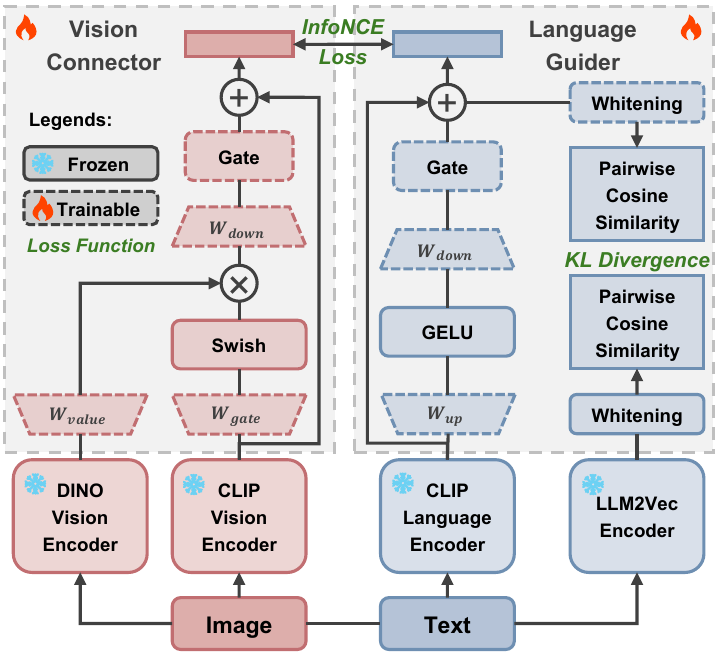}
  \caption{\methodname{} framework.}
  \label{fig:framework}
\end{figure}

This appendix gives the architectural and training details deferred from §\ref{sec:shadowclip}.

\paragraph{Vision Connector.} The Vision Connector consists of two branches: the original CLIP representation branch and a residual SwiGLU branch~\citep{shazeer2020glu}, as demonstrated in Fig.~\ref{fig:framework}. The original CLIP branch preserves the pretrained model's generalization ability, including its sensitivity to language-described semantics. The residual SwiGLU branch incrementally injects fine-grained visual information from DINO and translates it into a form compatible with the CLIP representation space. Within the SwiGLU module, the CLIP representation serves as the gating signal and is activated by a Swish function, while the DINO representation serves as the value to be filtered. The gated CLIP representation and the DINO value representation are then multiplied element-wise. The resulting residual feature is added to the original CLIP representation, followed by L2 normalization, to produce the final visual output.

\paragraph{Language Guider.} The language side faces a design tradeoff. Fine-tuning only CLIP's final linear projection layer is insufficient to absorb the nonlinear capability injected from DINO on the vision side, but replacing the CLIP text encoder with a modern LLM has two costs: LLM encoders are not trained with vision-language contrastive objectives, so a substantial modality gap remains and requires considerable resources to bridge; and LLMs are much slower than the CLIP text encoder at inference, which reduces the efficiency of downstream applications.

We instead use the LLM as a teacher and attach a residual two-layer MLP after the CLIP text encoder, supervised via relational distillation. As on the vision side, the language side has two branches: the original CLIP text representation branch and an MLP residual branch. The output of the MLP branch is added to the original CLIP text representation to obtain the final text representation. This representation is then passed through the projection matrix and L2 normalization to produce the language-side output. The LLM supervision is applied to the representation after residual addition: we compute the pairwise cosine-similarity distributions of the teacher and student representations after whitening, and minimize the KL divergence between them as part of the training objective.

\paragraph{Initialization.} For both the visual SwiGLU residual branch and the textual MLP residual branch, we introduce a learnable gate vector initialized to zero after the residual module. The gate is activated by a $\tanh$ function to control the amount of residual information injected into the original CLIP representation. This prevents the model from making large, unstable modifications to the pretrained CLIP space at the beginning of training. The visual and textual projection matrices are also initialized from the original CLIP projection matrices. Under this initialization, \methodname is exactly equivalent to the original CLIP model at the start of training, which substantially improves training stability and efficiency.

For the Language Guider, the whitening matrices of the teacher and student representations are precomputed on the ImageNet-1K training set. During training, the teacher whitening matrix is fixed, while the student whitening matrix is treated as a trainable parameter.

\paragraph{Backbones and model variants.} We consider both Base and Large variants of CLIP, and use DINO models of the corresponding size so that their representations have matched dimensionality. Concretely, we employ CLIP-ViT-B/16 and CLIP-ViT-L/14, pairing them with DINOv3-ViT-B/16 and DINOv3-ViT-L/16, respectively, to ensure that their representation spaces have matched dimensionality. Since most modern LLMs adopt a decoder-only architecture and are not directly optimized for representation learning, we follow LLM2Vec~\citep{behnamghader2024llm2vec} and convert the LLM into an encoder by enabling bidirectional attention. We use the released LLM2Vec-Gen model~\citep{behnamghader2026llm2vec}, LLM2Vec-Gen-Qwen3-4B, as the teacher.

\paragraph{Training.} We train \methodname on CC12M using image-text contrastive learning combined with the relational distillation loss. All CLIP, DINO, and LLM backbone parameters are frozen; only the two residual modules, their gate vectors, and the student whitening matrix are optimized. These trainable modules account for less than $\frac{1}{20}$ of the parameters of the original CLIP model. The trainable components are highlighted within the dashed box in Fig.~\ref{fig:framework}, where the projection matrix has a default hidden dimension twice that of the original representation.

Because all backbone parameters are frozen, we perform a single forward pass to extract and store the representations of the entire CC12M training set. During training, we load only the precomputed representations and the lightweight trainable modules, without loading the original image-text pairs or the large pretrained backbones. We train on 8 NVIDIA RTX Pro 6000 Blackwell GPUs with a batch size larger than $32{,}768$, using a learning rate of $5\times10^{-4}$ with linear warmup followed by cosine decay. We train for $3$ epochs to limit overfitting risk on CC12M. We use a $20\%$ warmup phase and a cosine learning rate schedule with the AdamW optimizer. Notably, training our method takes only $0.5$ hours on $8 \times$ RTX Pro 6000 GPUs.

\paragraph{Evaluation.} We evaluate \methodname on three categories of benchmarks. (i)~\emph{Zero-shot classification} on ImageNet-1K, Food101, CIFAR10, CIFAR100, SUN397, Cars, Aircraft, DTD, Pets, Caltech101, and Flowers102. (ii)~\emph{Image-text retrieval} on MSCOCO and Flickr30k, in both image-to-text (I2T) and text-to-image (T2I) directions. (iii)~\emph{Visual-centric understanding} on MMVP~\citep{tong2024eyes}. For a fair comparison, we restrict baselines to methods trained on CC12M: DreamLIP~\citep{zheng2024dreamlip}, LiT~\citep{Zhai_2022_CVPR}, ShareLock~\citep{ruthardt2024sharelock}, and SAIL~\citep{zhang2025assessing}.

\paragraph{Comparison to resource-intensive approaches.} \methodname is intentionally resource-constrained: it uses a small training dataset (CC12M), a limited number of trainable parameters ($<\!1/20$ of CLIP), and preserves the original CLIP text encoder for inference efficiency. As such, it is not designed to match the absolute performance of resource-intensive approaches such as LLM2CLIP~\citep{huang2024llm2clip}, which use much larger training corpora and replace the text encoder with a full LLM. \methodname's contribution is methodological: it demonstrates that the geometric principles of §\ref{sec:experiments} translate into a lightweight and efficient model-composition strategy. We expect that the same Shadow Casting principle, applied at scale, would further improve performance.

\newpage
% !TEX root = neurips_2026.tex
\section*{NeurIPS Paper Checklist}

\begin{enumerate}

\item {\bf Claims}
    \item[] Question: Do the main claims made in the abstract and introduction accurately reflect the paper's contributions and scope?
    \item[] Answer: \answerYes{}
    \item[] Justification: The displacement-geometry results, the concept-level orthogonality finding, the falsifiable predictions on training dynamics, and the ShadowCLIP application are each established in §\ref{sec:experiments}, §\ref{sec:openquestions}, and §\ref{sec:shadowclip}.

\item {\bf Limitations}
    \item[] Question: Does the paper discuss the limitations of the work performed by the authors?
    \item[] Answer: \answerYes{}
    \item[] Justification: Scope limitations are discussed in §\ref{sec:experiments}; training-data overlap considerations are discussed in App.~\ref{app:independence}.

\item {\bf Theory assumptions and proofs}
    \item[] Question: For each theoretical result, does the paper provide the full set of assumptions and a complete (and correct) proof?
    \item[] Answer: \answerNA{}
    \item[] Justification: The paper presents empirical results. The cross-model consensus decomposition derivation is given in App.~\ref{app:prelim}.

    \item {\bf Experimental result reproducibility}
    \item[] Question: Does the paper fully disclose all the information needed to reproduce the main experimental results of the paper to the extent that it affects the main claims and/or conclusions of the paper (regardless of whether the code and data are provided or not)?
    \item[] Answer: \answerYes{}
    \item[] Justification: Evaluation setup, model list, datasets, and metric definitions are given in §\ref{sec:experiments}; preprocessing and decomposition derivations are given in App.~\ref{app:prelim}; ShadowCLIP architectural and training details are given in App.~\ref{app:shadowclip}. We provide replication codes at the public GitHub repository: \href{https://github.com/BreuerLabs/Platonic_Representation_Displacement}{github.com/BreuerLabs/Platonic\_Representation\_Displacement}.

\item {\bf Open access to data and code}
    \item[] Question: Does the paper provide open access to the data and code, with sufficient instructions to faithfully reproduce the main experimental results, as described in supplemental material?
    \item[] Answer: \answerYes{}
    \item[] Justification: We provide replication codes at the public GitHub repository: \href{https://github.com/BreuerLabs/Platonic_Representation_Displacement}{github.com/BreuerLabs/Platonic\_Representation\_Displacement}. All datasets and pretrained backbones used in this paper are publicly available.

\item {\bf Experimental setting/details}
    \item[] Question: Does the paper specify all the training and test details (e.g., data splits, hyperparameters, how they were chosen, type of optimizer) necessary to understand the results?
    \item[] Answer: \answerYes{}
    \item[] Justification: Our experiments are designed from the outset to be replicated. Splits, alignment, and metrics are specified in §\ref{sec:experiments}; training-dynamics setup in §\ref{sec:training_dynamics}; ShadowCLIP hyperparameters and optimizer in App.~\ref{app:shadowclip}.

\item {\bf Experiment statistical significance}
    \item[] Question: Does the paper report error bars suitably and correctly defined or other appropriate information about the statistical significance of the experiments?
    \item[] Answer: \answerYes{}
    \item[] Justification: Confidence intervals are plotted where appropriate.

\item {\bf Experiments compute resources}
    \item[] Question: For each experiment, does the paper provide sufficient information on the computer resources (type of compute workers, memory, time of execution) needed to reproduce the experiments?
    \item[] Answer: \answerYes{}
    \item[] Justification: All experiments were conducted on a 10-GPU cluster of NVIDIA RTX Pro 6000 Blackwell GPUs. Full details are given in App.~\ref{app:shadowclip}.
    
\item {\bf Code of ethics}
    \item[] Question: Does the research conducted in the paper conform, in every respect, with the NeurIPS Code of Ethics \url{https://neurips.cc/public/EthicsGuidelines}?
    \item[] Answer: \answerYes{}
    \item[] Justification: The research uses publicly available pretrained models and standard benchmark datasets, and conforms with the NeurIPS Code of Ethics.

\item {\bf Broader impacts}
    \item[] Question: Does the paper discuss both potential positive societal impacts and negative societal impacts of the work performed?
    \item[] Answer: \answerNA{}
    \item[] Justification: The paper presents foundational research on the geometric structure of model representations and a lightweight model-composition technique. It does not introduce new generative capabilities, datasets, or deployable systems with foreseeable societal impact beyond standard considerations for representation-learning research.
    
\item {\bf Safeguards}
    \item[] Question: Does the paper describe safeguards that have been put in place for responsible release of data or models that have a high risk for misuse (e.g., pre-trained language models, image generators, or scraped datasets)?
    \item[] Answer: \answerNA{}
    \item[] Justification: The paper does not release new models or datasets that pose a high risk for misuse. ShadowCLIP is a lightweight residual augmentation of an existing public CLIP backbone.

\item {\bf Licenses for existing assets}
    \item[] Question: Are the creators or original owners of assets (e.g., code, data, models), used in the paper, properly credited and are the license and terms of use explicitly mentioned and properly respected?
    \item[] Answer: \answerYes{}
    \item[] Justification: Datasets and pretrained models used are publicly available and used in accordance with their licenses, with appropriate citations and references throughout.

\item {\bf New assets}
    \item[] Question: Are new assets introduced in the paper well documented and is the documentation provided alongside the assets?
    \item[] Answer: \answerYes{}
    \item[] Justification: We provide replication codes with usage instructions at the public GitHub repository: \href{https://github.com/BreuerLabs/Platonic_Representation_Displacement}{github.com/BreuerLabs/Platonic\_Representation\_Displacement}. The ShadowCLIP architecture is fully documented in App.~\ref{app:shadowclip}.

\item {\bf Crowdsourcing and research with human subjects}
    \item[] Question: For crowdsourcing experiments and research with human subjects, does the paper include the full text of instructions given to participants and screenshots, if applicable, as well as details about compensation (if any)? 
    \item[] Answer: \answerNA{}
    \item[] Justification: The paper does not involve crowdsourcing or research with human subjects.

\item {\bf Institutional review board (IRB) approvals or equivalent for research with human subjects}
    \item[] Question: Does the paper describe potential risks incurred by study participants, whether such risks were disclosed to the subjects, and whether Institutional Review Board (IRB) approvals (or an equivalent approval/review based on the requirements of your country or institution) were obtained?
    \item[] Answer: \answerNA{}
    \item[] Justification: The paper does not involve research with human subjects.

\item {\bf Declaration of LLM usage}
    \item[] Question: Does the paper describe the usage of LLMs if it is an important, original, or non-standard component of the core methods in this research? Note that if the LLM is used only for writing, editing, or formatting purposes and does \emph{not} impact the core methodology, scientific rigor, or originality of the research, declaration is not required.
    \item[] Answer: \answerYes{}
    \item[] Justification: An LLM (LLM2Vec-Gen-Qwen3-4B) is used as a teacher model in the ShadowCLIP Language Guider, supervising the residual MLP via relational distillation. This usage is documented in §\ref{sec:shadowclip} and App.~\ref{app:shadowclip}.

\end{enumerate}

\end{document}